\documentclass[times,twocolumn,a4paper,10pt]{article}
\usepackage{abstract}

\usepackage{etoolbox}
\patchcmd\thebibliography
 {\labelsep}
 {\labelsep\itemsep=0pt\relax}
 {}
 {\typeout{Couldn't patch the command}}

\usepackage{geometry}
\usepackage{framed,multirow,multicol}

\usepackage{amssymb}
\usepackage{latexsym}

\usepackage{url}
\usepackage{xcolor}
\definecolor{newcolor}{rgb}{.8,.349,.1}
\usepackage{xspace}

\usepackage{amsmath,amsfonts,amssymb}
\usepackage[lined,noend,linesnumbered]{algorithm2e}
\usepackage[capitalize]{cleveref}
\usepackage{float}

\usepackage{graphicx}
\usepackage{caption}
\usepackage{subcaption}
\graphicspath{{figures/}}

\newcommand{\ie}{\textit{i.e.}\xspace}
\newcommand{\eg}{\textit{e.g.}\xspace}

\newcommand{\resp}{\textit{resp.}\xspace}
\newcommand{\Attribute}{\ensuremath{\mathcal{A}}\xspace}
\newcommand{\Tree}{\ensuremath{\mathcal{T}}\xspace}
\newcommand{\Chi}{\ensuremath{\mathcal{X}}\xspace}

\newcommand{\extinct}{\ensuremath{\mathcal{E}}\xspace}
\newcommand{\map}{\ensuremath{\mathcal{M}}\xspace}

\newenvironment{bluetext}{\color{black}}{\ignorespacesafterend}

\definecolor{forestgreen}{RGB}{0,210,0}

\usepackage{etoolbox}
\AtBeginEnvironment{algorithm}{\linespread{0.5}\selectfont}
\SetAlCapSkip{2ex}

\begin{document}

\thispagestyle{empty}

\title{Hierarchical image simplification and segmentation based on
  Mumford-Shah-salient level line selection}

\author{\small{Yongchao Xu\thanks{EPITA Research and Development
      Laboratory (LRDE), 14-16, rue Voltaire, FR-94276 Le
      Kremlin-Bic\^etre, France}~\thanks{Department of Signal and
      Image Processing, Telecom ParisTech, 46 rue Barrault, 75013
      Paris, France}, Thierry {G\'eraud}\footnotemark[1], Laurent
    {Najman}\thanks{Universit\'e Paris-Est, Laboratoire d'Informatique
      Gaspard-Monge (LIGM), A3SI, ESIEE Paris, Cit\'e Descartes, BP
      99, FR-93162 Noisy-le-Grand, France\newline E-mails:
      \texttt{\{yongchao.xu, thierry.geraud\}@lrde.epita.fr,
        l.najman@esiee.fr}\newline The paper is accepted in
      Pattern Recognition Letters.  }}}

\date{}


\newcommand\makeAbstract{%
\begin{center}\textbf{Abstract}\end{center}
\begin{list}{}{\leftmargin=3em\rightmargin=\leftmargin}\item\relax
\small
Hierarchies, such as the tree of shapes, are popular representations
for image simplification and segmentation thanks to their multiscale
structures. Selecting meaningful level lines (boundaries of shapes)
yields to simplify image while preserving intact salient
structures. Many image simplification and segmentation methods are
driven by the optimization of an energy functional, for instance the
celebrated Mumford-Shah functional. In this paper, we propose an
efficient approach to hierarchical image simplification and
segmentation based on the minimization of {\bluetext the
  piecewise-constant Mumford-Shah} functional.  This method conforms
to the current trend that consists in producing hierarchical results
rather than a unique partition. Contrary to classical approaches which
compute optimal hierarchical segmentations from an input hierarchy of
segmentations, we rely on the tree of shapes, a unique and
well-defined representation equivalent to the image. Simply put, we
compute for each level line of the image an attribute function that
characterizes its persistence under the energy minimization. Then we
stack the level lines from meaningless ones to salient ones through a
saliency map based on extinction values defined on the tree-based
shape space.  Qualitative illustrations and quantitative evaluation on
Weizmann segmentation evaluation database demonstrate the
state-of-the-art performance of our method.
\end{list}\par\vspace{6mm}%
}

\twocolumn[\maketitle\makeAbstract]

\saythanks






\section{Introduction}
\label{sec:introduction}

In natural images, meaningful contours are usually smooth and
well-contrasted. Many authors (\eg,~\cite{caselles.99.ijcv,
  cao.05.jmiv}) claim that significant contours of objects in images
coincide with segments of the image level lines. The level lines are
the boundaries of the connected components described by the {\em tree
  of shapes} proposed in~\cite{monasse.00.itip}, and also known as
{\em topographic map} in~\cite{caselles.99.ijcv}. Image simplification
or segmentation can then be obtained by selecting meaningful level
lines in that tree. This subject has been investigated in the past
by~\cite{pardo.02.icip, cao.05.jmiv,
  cardelino.06.icip}. In~\cite{lu.07.itip}, the authors have proposed
a tree simplification method for image simplification purpose based on
the binary partition tree.

Classically, finding relevant contours is often tackled using an
energy-based approach. It involves minimizing a two-term-based energy
functional of the form $E_{\lambda_s} = \lambda_s C + D$, where $C$ is
the regularization term controlling the regularity of contours, $D$ is
a data fidelity term, and $\lambda_s$ is a parameter. A popular
example is the seminal work of~\cite{mumford.89.cpam}. Curve evolution
methods are usually used to solve this minimization problem. They have
solid theoretical foundations, yet they are often computational
expensive.

Current trends in image simplification and segmentation are to find a
multiscale representation of the image rather than a unique
partition. There exist many works about hierarchical segmentations
such as the geodesic saliency of watershed contours proposed
in~\cite{najman.96.pami} and gpb-owt-ucm proposed
by~\cite{arbelaez.11.pami} and references therein. Some authors
propose to minimize a two-term-based energy functional subordinated to
a given input hierarchy of segmentations, in order to find an optimal
hierarchical image segmentations in the sense of energy
minimization. Examples are the works of~\cite{guigues.06.ijcv,
  kiran.14.pr}. Yet, the choice or the construction of the input
hierarchy of segmentations for these methods is an interesting problem
in itself. \cite{perret.15.ismm} compared different choices of
morphological hierarchies for supervised segmentation.

In this paper we propose a novel hierarchical image simplification and
segmentation based on minimization of an energy functional (\eg, the
piecewise-constant Mumford-shah functional). The minimization is
performed subordinated to the shape space given by the tree of shapes,
a unique and equivalent image representation. The basis of our
proposal was exposed in our previous study in~\cite{xu.13.icip}, in
which we proposed an efficient greedy algorithm computing a locally
optimal solution of the energy minimization problem. The basic idea is
to take into account the meaningfulness of each level line {\bluetext
  which measures its ``importance''. An example of meaningfulness
  function that we will use through the paper is the average of
  gradient's magnitude along level lines. The order based on these
  meaningfulness values allows to get very quickly a locally optimal
  solution, which yields a well-simplified image while preserving the
  salient structures.} The current paper extends this idea to
hierarchical simplification and segmentation. More precisely,
following the same principle but without fixing the parameter
$\lambda_s$ in the two-term-based energy, we compute an attribute
function that characterizes the persistence of each shape under the
energy minimization. Then we compute a saliency map, a single image
representing the complete hierarchical simplifications or
segmentations. To do so, we rely on the idea of hierarchy
transformation via extinction value proposed by~\cite{vachier.95.nlsp}
and on the framework of tree-based shape space introduced
in~\cite{xu.14.filter}. This scheme of hierarchy transformation has
been first used in~\cite{xu.13.ismm} for a different input hierarchy
and attribute function. Related algorithms were presented
in~\cite{xu.15.ismm}. The present paper extends on these ideas,
focusing on the computation of an attribute function related to energy
minimization.

The main contribution of this current paper is {\bluetext the
  proposition of a general framework of} hierarchical image
simplification and segmentation method based on energy minimization
subordinated to the tree of shapes, {\bluetext contrary to the
  classical approaches that are subordinated to an initial hierarchy
  of segmentations. It is based on the introduction of a novel
  attribute function $\Attribute_{\lambda_s}$ related to energy
  minimization. } We have tested {\bluetext the proposed framework}
with a very simple segmentation model in this paper. Despite its
simplicity, we obtain results that are competitive with the ones of
some state-of-the-art methods on the classical segmentation dataset
from~\cite{alpert.12.pami}. In particular, they are on par with
Gpb-owt-ucm proposed in~\cite{arbelaez.11.pami} on this dataset.


The rest of this paper is organized as follows: Some background
information is provided in
Section~\ref{sec:background}. Section~\ref{sec:method} is dedicated to
depict the proposed method, followed by some illustrations and
experimental results in
Section~\ref{sec:illustration}. Section~\ref{sec:relatedwork} compares
the proposed method with some similar works. We then conclude in
Section~\ref{sec:conclusion}.


\section{Background}
\label{sec:background}

\subsection{The Tree of shapes}
\label{subsec:tos}

For any $\lambda \in \mathbb{R} \textrm{ or } \mathbb{Z}$, the upper
level sets $\Chi_\lambda$ and lower level sets $\Chi^\lambda$ of an
image $f: \Omega \rightarrow \mathbb{R} \textrm{ or } \mathbb{Z}$ are
respectively defined by $\Chi_\lambda(f) = \{ p \in \Omega \mid f(p)
\ge \lambda \}$ and $\Chi^\lambda(f) = \{ p \in \Omega \mid f(p) \le
\lambda \}.$ Both upper and lower level sets have a natural inclusion
structure: $\forall \, \lambda_1 \leq \lambda_2, \; \Chi_{\lambda_1}
\supseteq \Chi_{\lambda_2} \,\mbox{~and~}\, \Chi^{\lambda_1} \subseteq
\Chi^{\lambda_2},$ which leads to two distinct and dual
representations of an image, the max-tree and the min-tree.

\begin{figure}
  \centering \includegraphics[width=0.8\linewidth]{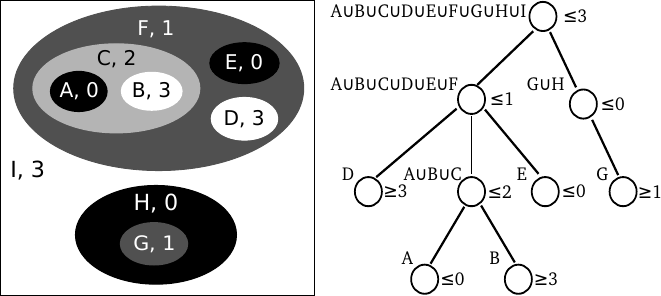}
  \caption{An image (left) and its tree of shapes (right).}
  \label{fig:tree}
\end{figure}

Another tree has been introduced in~\cite{monasse.00.itip} via the
notion of shapes. A \emph{shape} is defined as a connected component
of an upper or lower level set where its holes have been filled
in. Thanks to the inclusion relationship of both kinds of level sets,
the set of shapes gives a unique tree, called \emph{tree of shapes}.
This tree is a self-dual, non-redundant, and complete representation
of an image. It is equivalent to the input image in the sense that the
image can be reconstructed from the tree. And it is invariant to
affine contrast changes. Such a tree also inherently embeds a
morphological scale-space (the parent of a node/shape is a larger
shape).  An example on a synthetic image is depicted in
Fig.~\ref{fig:tree}. Recently, an extension of the tree of shapes for
color images has been proposed by~\cite{carlinet.15.itip} through the
inclusion relationship between the shapes of its three grayscale
channels.


\subsection{Hierarchy of image segmentations or saliency maps}
\label{subsec:hos}

A hierarchy of image segmentation $H$ is a multiscale representation
that consists of a set of nesting partitions from fine to coarse: $H =
\{\mathcal{P}_i \, | \, 0 \leq i \leq n, \forall j, k, \, 0 \leq j
\leq k \leq n \Rightarrow \mathcal{P}_j \sqsubseteq \mathcal{P}_k\},$
where $\mathcal{P}_n$ is the partition $\{\Omega\}$ of $\Omega$ into a
single region, and $\mathcal{P}_0$ represents the finest partition of
the image $f$. $\mathcal{P}_j \sqsubseteq \mathcal{P}_k$ implies that
the partition $\mathcal{P}_j$ is finer than $\mathcal{P}_k$, which
means $\forall \, R \in \mathcal{P}_j, \exists \, R' \in \mathcal{P}_k$
such that $R \subseteq R'$.

As a multiscale representation, a hierarchy of segmentation satisfies
the most fundamental principle for multiscale analysis: the causality
principle presented by~\cite{koenderink.84.bc}. From this principle,
for any couple of scales $\lambda_{s_2} > \lambda_{s_1}$, the
``structures'' found at scale $\lambda_{s_2}$ should find a ``cause''
at scale $\lambda_{s_1}$. In the case of a hierarchy of segmentation,
following the work of~\cite{guigues.06.ijcv}, the causality principle
is applied to the edges associated to the set of partitions spanned by
$H$: for any pair of scales $\lambda_{s_2} > \lambda_{s_1}$, the
boundaries of partition $\mathcal{P}_{\lambda_{s_2}}$ are in a
one-to-one mapping with a subset of the boundaries of
$\mathcal{P}_{\lambda_{s_1}}$ (their ``cause''). The pair $(H,
\lambda_s)$ is called an indexed hierarchy.

A useful representation of hierarchical image segmentations was
originally introduced in~\cite{najman.96.pami} under the name of {\em
  saliency map}. A saliency map is obtained by stacking a family of
hierarchical contours. This representation was then rediscovered
independently by~\cite{guigues.06.ijcv} through the notion of
scale-set theory for visualization purposes, and it is then
popularized by~\cite{arbelaez.11.pami} under the name of {\em
  ultrametric contour map} for boundary extraction and comparing
hierarchies. It has been proved theoretically in~\cite{najman.11.jmiv}
that a hierarchy of segmentations is equivalent to a saliency map.
Roughly speaking, for a given indexed hierarchy $(H, \lambda)$, the
corresponding saliency map can be obtained by weighing each contour of
the image domain with the highest value $\lambda_s$ such that it
appears in the boundaries of some partition represented by the
hierarchy $H$. The low level (resp. upper level) of a hierarchy
corresponds to weak (resp. strong) contours, and thus an
over-segmentation (resp. under-segmentation) can be obtained by
thresholding the saliency map with low (resp. high) value.


\subsection{From shape-space filtering to hierarchy of segmentations}
\label{subsec:ssf}

The three morphological trees reviewed in Section~\ref{subsec:tos} and
the hierarchies of segmentations reviewed in Section~\ref{subsec:hos}
have a tree structure. Each representation is composed of a set of
connected components $\mathbb{C}$. Any two different elements $C_i \in
\mathbb{C}, C_j \in \mathbb{C}$ are either disjoint or nested:
$\forall \, C_i \in \mathbb{C}, \; C_j \in \mathbb{C}, C_i \, \cap \,
C_j \neq \emptyset \, \Rightarrow \, C_i \subseteq C_j \textrm{ or }
C_j \subseteq C_i$. This property leads to the definition of {\em
  tree-based shape space} in~\cite{xu.14.filter}: a graph
representation $G_{\mathbb{C}} = (\mathbb{C}, E_{\mathbb{C}})$, where
each node of the graph represents a connected component in the tree,
and the edges $E_{\mathbb{C}}$ are given by the inclusion relationship
between connected components in $\mathbb{C}$. In~\cite{xu.14.filter},
we have proposed to filter this shape space by applying some classical
operators, notably connected operators on $G_{\mathbb{C}}$. We have
shown that this shape-space filtering encompasses some classical
connected operators, and introduces two families of novel connected
operators: shape-based lower/upper leveling and shaping.

\begin{figure}
  \begin{center}
    \begin{minipage}[b]{0.7\linewidth}
      \centerline{\includegraphics[width=0.6\linewidth]{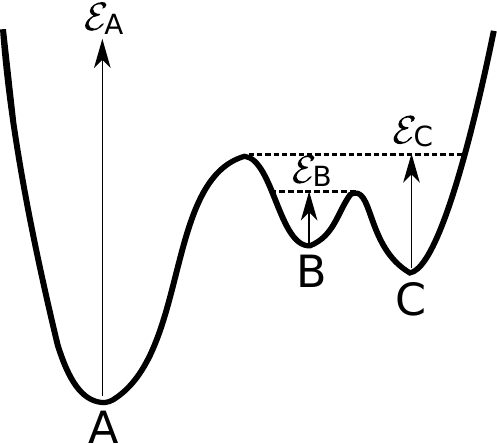}}
    \end{minipage}
  \end{center}
\caption{Illustration of the extinction values $\extinct$ of three
  minima. The order is $A \prec C \prec B$. $B$ merges with $C$, $C$
  merges with $A$.}
\label{fig:extinction}
\end{figure}

Instead of filtering the shape space, another idea is to consider each
region of the shape space as a candidate region of a final
partition. For example, we weigh the shape space by a quantitative
{\bluetext attribute} $\Attribute$. Then each local minimum of the
node-weighted shape space is considered as a candidate region of a
partition. The importance of each local minimum (\ie, each region) can
be measured quantitatively by the extinction value $\extinct$ proposed
by~\cite{vachier.95.nlsp}. Let $\prec$ be a strict total order on the
set of minima $m_1\prec m_2 \prec\ldots$, such that $m_i\prec m_{i+1}$
whenever $\Attribute(m_i) < \Attribute(m_{i+1})$. Let $CC$ be the
lowest lower level connected component (defined on the shape space)
that contains both $m_{i+1}$ and a minimum $m_j$ with $j<(i+1)$. The
extinction value for the minimum $m_{i+1}$ is defined as the
difference of level of $CC$ and $\Attribute(m_{i+1})$.  An example of
extinction values for three minima is depicted in
Fig~\ref{fig:extinction}. We weigh the boundaries of the regions
corresponding to the local minima with the extinction values. This
yields a saliency map representing a hierarchical image simplification
or segmentation. This scheme allows to transform any hierarchical
representation into a hierarchical segmentation. It has been firstly
used in~\cite{xu.13.ismm}, where the input hierarchy is a minimum
spanning tree and the attribute is computed locally inspired from the
work of~\cite{felzenszwalb.04.ijcv}.  For the sake of completeness,
the algorithm~\cite{xu.15.ismm} for the extinction-based hierarchy
transformation is presented in Section~\ref{subsec:algorithm}.

\subsection{Energy-based simplification and segmentation}
\label{subsec:mumford}
There exist several works of hierarchical image segmentations based on
energy minimization (see~\cite{guigues.06.ijcv}
and~\cite{kiran.14.pr}). A general formulation of these methods
involves minimizing a two term-based energy functional of the form
$E_{\lambda_s} = \lambda_s C + D$. $C$ is a regularization term, $D$
is data fidelity term, and $\lambda_s$ is a parameter. Let $ \{R\} =
R_1 \sqcup \dots \sqcup R_n$ be a partition of the image domain. If
the energy functional can be written by $E_{\lambda_s} = \sum_{R_i \in
  \{R\}} \big(\lambda_s C(R_i) + D(R_i)\big)$, then $E_{\lambda_s}$ is
called an affine separable energy functional. Furthermore, if either
the regularization term $C$ decreases or the data fidelity term $D$
increases, the energy $E_{\lambda_s}$ is multiscale affine
separable. A popular instance of such an energy functional that we
will use as an example through this paper is the piecewise-constant
Mumford-Shah functional proposed by~\cite{mumford.89.cpam}. For an
image $f$, it is given by
\begin{equation}
  E_{\lambda_s}(f, \partial  \{R\}) \,= \iint_{\{R\}} \!\! (\tilde{f}_i - f)^2 \, dx dy \,+\,
  \lambda_s \, |\partial \{R\}|,
\label{eq:simplemumford}
\end{equation}
where $\tilde{f}_i = \frac{1}{|R_i|} \iint_{R_i} f \, dx dy$ inside
each region $R_i \in \{R\}$, $\partial \{R\}$ is the set of contour,
and $|\cdot|$ denotes the cardinality.
%



\section{Hierarchical image simplification and segmentation via level line selection}
\label{sec:method}

\subsection{Main idea}
\label{subsec:mainidea}

The current proposal {\bluetext is a general framework of hierarchical
  image simplification and segmentation based on energy minimization
  subordinated to the tree of shapes. It extends} a preliminary
version of this study in~\cite{xu.13.icip} {\bluetext that selects}
salient level lines based on Mumford-Shah energy functional
minimization. We review this non-hierarchical version in
Section~\ref{subsec:llselection}, using a more general multiscale
affine separable energy.
{\bluetext The hierarchical version proposed in the current paper is
  achieved thanks to:}
\begin{itemize}
\item the introduction of a novel attribute function
  $\Attribute_{\lambda_s}$ for each level line related to the
energy regularization parameter $\lambda_s$,
\item and the idea of
hierarchy transformation based on extinction values and on a
tree-based shape space.
\end{itemize}
This is detailed in
Section~\ref{subsec:hslls}. Section~\ref{subsec:algorithm} provides
algorithms for the whole process.

\subsection{Image simplification by salient level line selection}
\label{subsec:llselection}

For a given tree of shapes $\Tree$ composed of a set of shapes
$\{\tau_i\}$, any two successive shapes of $\Tree$ are related by an
edge reflecting the inclusion relationship, also known as the
parenthood between nodes of the tree. This tree structure $\Tree$
provides an associated partition of the image $\{R_\Tree\} =
R_{\tau_1} \sqcup \dots \sqcup R_{\tau_n}$, where $R_\tau = \{p \, |
\, p \in \tau, p \notin \mathit{Ch}(\tau) \}$ with $\mathit{Ch}(\tau)$
representing all the children of the shape $\tau$. We denote by
$E_{\lambda_s}(f, \Tree)$ the energy functional (see
Section~\ref{subsec:mumford}) subordinated to the tree by considering
its associated partition $\{R_\Tree\}$. This energy minimization is
given by:
\begin{equation}
  \underset{\Tree'}{\min} \; E_{\lambda_s}(f, \Tree'),
  \label{eq:solution}
\end{equation}
where $\Tree'$ is a simplified version of $\Tree$ by removing some
shapes from $\Tree$ and by updating the parenthood relationship.

\begin{figure}
  \centering
  \includegraphics[width=0.8\linewidth]{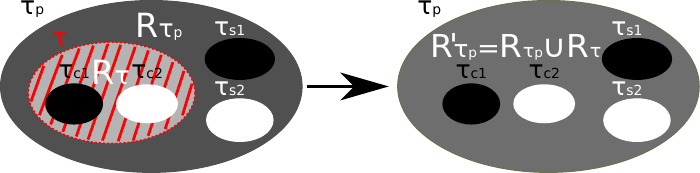}
  \caption{Suppressing the node $\tau$ makes the ``region'' $R_\tau$
    (covered with red oblique lines) merge with $R_{\tau_p}$; the
    result (depicted in the right image) is a simplified image.}
  \label{fig:merging}
\end{figure}

The basic operation of the energy minimization problem given by
Eq.~\eqref{eq:solution} is to remove the contours of some
  shapes $\{\tau\}$ included in their corresponding parents
  $\{\tau_p\}$,
which triggers the update of $R'_{\tau_p} \, = \, R_{\tau_p} \cup
R_\tau$ for each shape $\tau$. The parent of its children $\tau_{c1},
\dots , \tau_{ck}$ should also be updated to the
$\tau_p$. Fig.~\ref{fig:merging} shows an example of a such merging
operation.

Observe that the minimization problem of Eq.~(\ref{eq:solution}) is a
combinatorial optimization. The computation of the optimum has an
exponential complexity. Hence a greedy algorithm is usually applied to
compute a local optimum instead of a global optimum (see
  also~\cite{ballester.07.jmiv}). It iteratively removes
  the shapes to decrease the energy functional. The greedy algorithm
stops when no other shape can be removed that favors a decrease of the
energy. The removability of a shape $\tau$ is decided by the sign of
the energy variation $\Delta E_{\lambda_s}^\tau$ while $\tau$ is
suppressed. For the multiscale affine separable energy described in
Section~\ref{subsec:mumford}, $\Delta E_{\lambda_s}^\tau$ is given by:
\begin{equation}
  \Delta E_{\lambda_s}^\tau = D(R'_{\tau_p}) - D(R_{\tau}) -
  D(R_{\tau_p}) - \lambda_s\big(C(R_{\tau}) + C(R_{\tau_p}) - C(R'_{\tau_p})\big).
  \label{eq:variation}
\end{equation}
Taking the piecewise-constant Mumford-Shah functional given by
Eq.~\eqref{eq:simplemumford} as an energy example, and let $S(f, R_i)$
be the sum of value of all the pixels inside $R_i$, Then the
functional variation $\Delta E_{\lambda_s}^\tau$ is given by:
\begin{equation}
  \Delta E_{\lambda_s}^\tau = \frac{S^2(f,{R_\tau})}{|R_\tau|} +
  \frac{S^2(f, R_{\tau_p})}{|R_{\tau_p}|} - \frac{S^2(f,
    R'_{\tau_p})}{|R'_{\tau_p}|} - \lambda_s |\partial
  \tau|.
  \label{eq:variationmumford}
\end{equation}

If $\Delta E_{\lambda_s}^\tau$ is negative, which means the
suppression of $\tau$ decreases the functional, then we remove
$\tau$. According to Eq.~\eqref{eq:variation}, the removability of a
shape $\tau$ depends only on $R_\tau$ and $R_{\tau_p}$. As the
suppression of the shape $\tau$ triggers the update of $R_{\tau_p}$,
the removal of $\tau$ impacts also the removability of its parent, its
children and siblings. So the order of level line removal is
critical. In~\cite{xu.13.icip}, we proposed to fix the order by
sorting the level lines in increasing order of a
  quantitative meaningfulness attribute $\Attribute$ (\eg, the average of
gradient's magnitude along the level line $\Attribute_\nabla$).


Meaningful contours in natural images are usually well-contrasted and
smooth. Indeed, the minimization of energy functional in
Eq~(\ref{eq:simplemumford}) favors the removal of level lines having
small contrast (by data fidelity term) or being complex (by
regularization term). So the shapes having small (\resp great)
attribute $\Attribute_\nabla$ are easier (\resp more difficult) to
filter out under the energy minimization process of
Eq.~\eqref{eq:solution}. Consequently, the level line sorting based on
attribute $\Attribute_\nabla$ provides a reasonable order to perform
the level lines suppression that makes the energy functional decrease.
Indeed, initially, each region $R_\tau$ has only several pixels. At
the beginning, many ``meaningless'' regions are removed, which forms
more proper regions in Eq.~\eqref{eq:variation} for the ``meaningful''
regions. The removal decisions for these ``meaningful'' regions based
on the sign of Eq.~\eqref{eq:variation} are more robust.


\subsection{Hierarchical salient level line selection}
\label{subsec:hslls}

\begin{figure}
  \centering
  \begin{subfigure}{0.31\linewidth}
    \centering
    \includegraphics[width=1.0\linewidth]{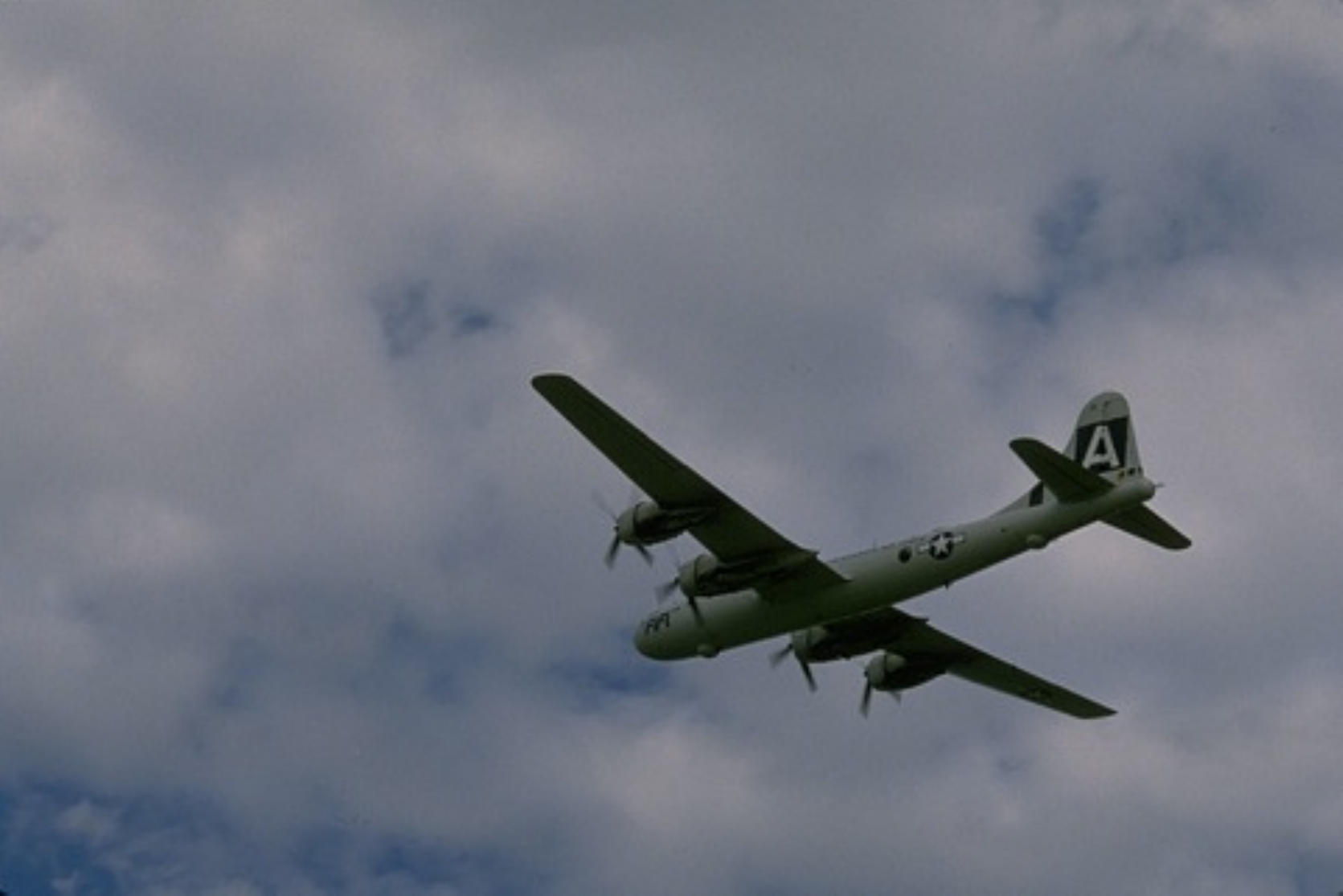}
  \end{subfigure}
  ~
  \begin{subfigure}{0.31\linewidth}
    \centering
    \includegraphics[width=1.0\linewidth]{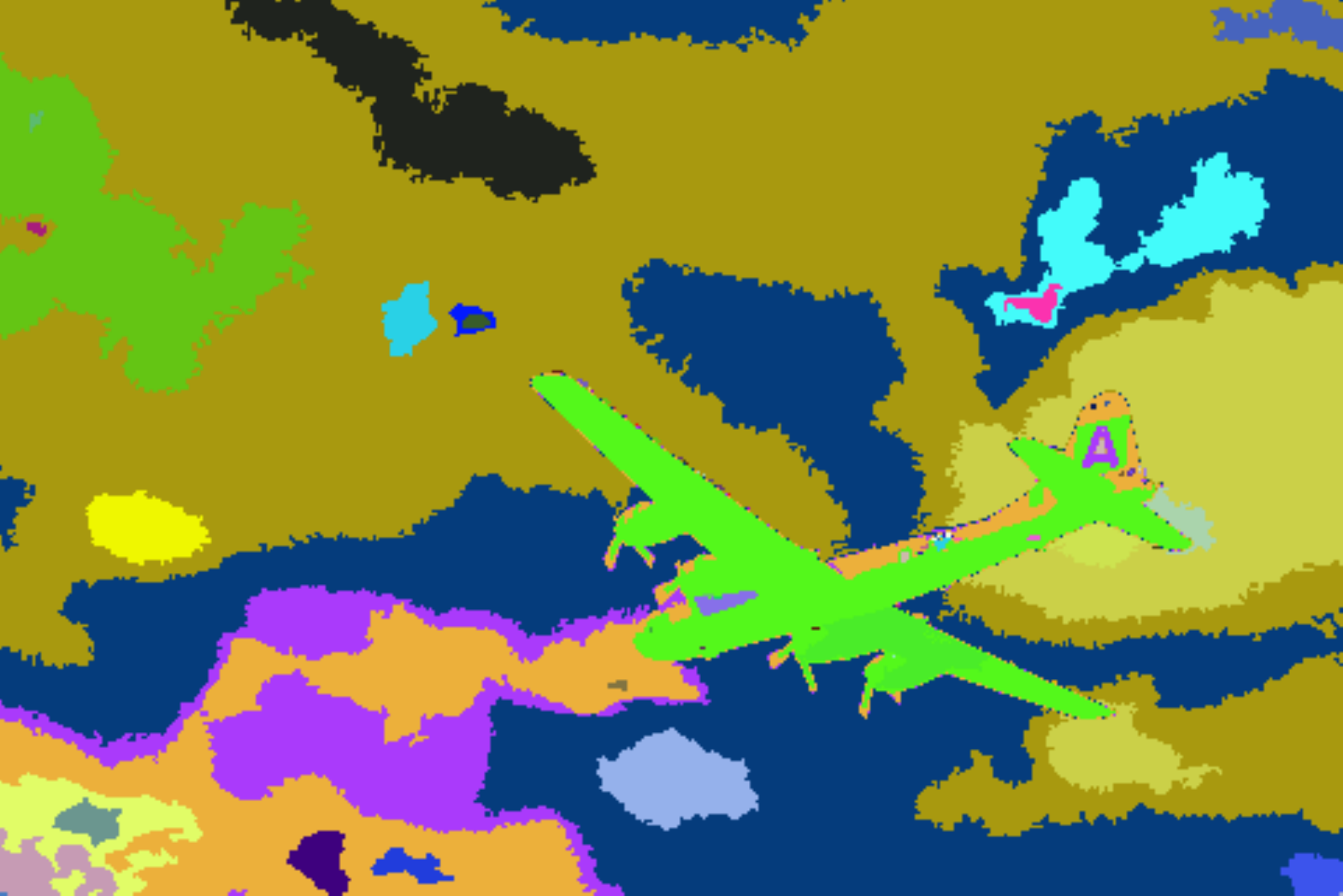}
  \end{subfigure}
  ~
  \begin{subfigure}{0.31\linewidth}
    \centering
    \includegraphics[width=1.0\linewidth]{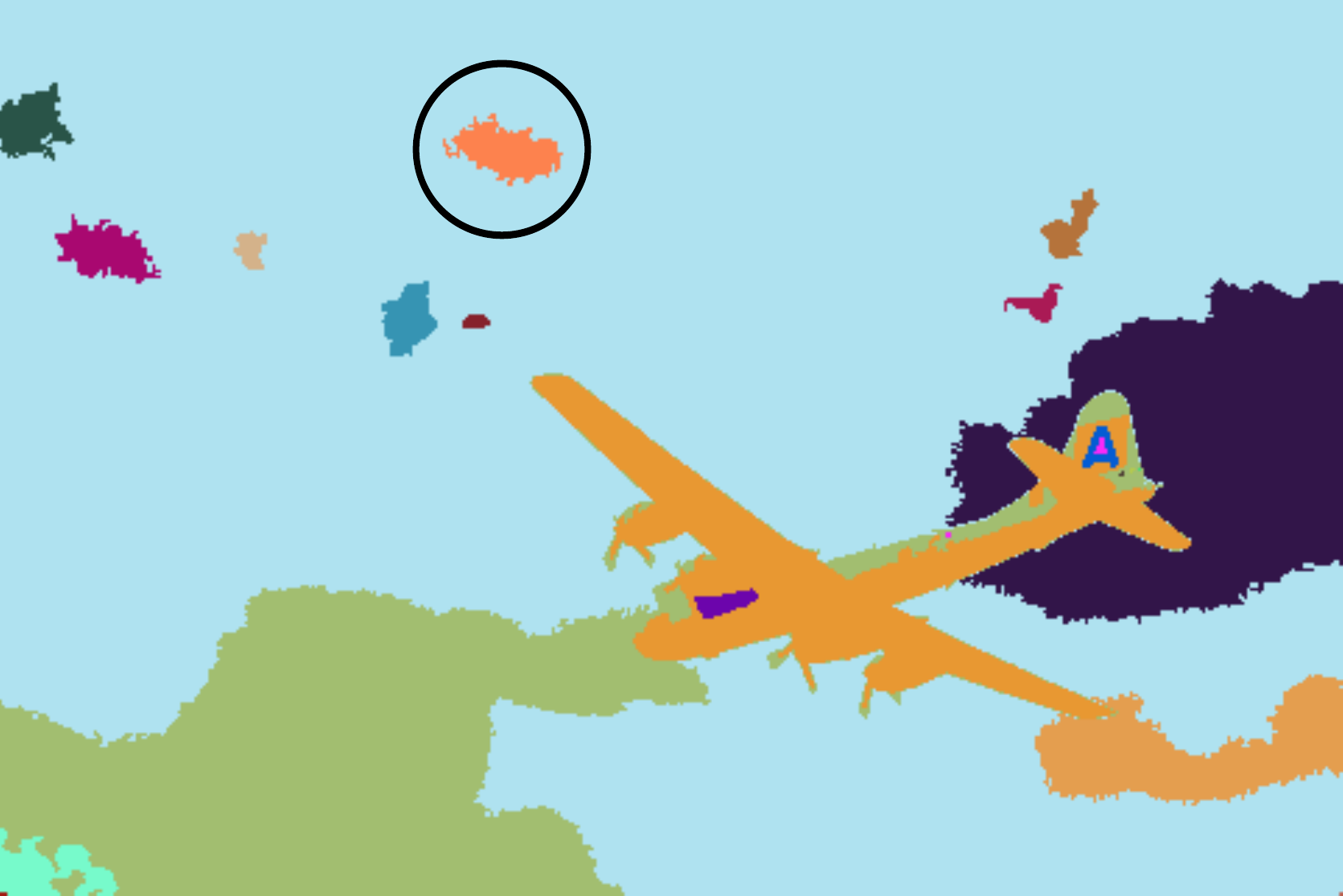}
  \end{subfigure}
  \caption{Illustration of causality principle
    violation. Left: input image; Middle and right: randomly colorized
    simplified image with $\lambda_{s_1} = 100$ and respectively
    $\lambda_{s_2} = 500$. The orange shape on top middle of the right
    image (surrounded by a black circle) is preserved for
    $\lambda_{s_2} = 500$, while it is removed for $\lambda_{s_1} =
    100$ in the middle image.}
  \label{fig:causality}
\end{figure}

The parameter $\lambda_s$, in the multiscale affine separable energy,
controls the simplification/segmentation degree for the method
described in Section~\ref{subsec:llselection}, which is however not
hierarchical. Because some level line $\tau$ may be removed with a
parameter $\lambda_{s_1}$, but preserved for a bigger $\lambda_{s_2} >
\lambda_{s_1}$. This contradicts the causality principle for
hierarchical image simplification/segmentation described in
Section~\ref{subsec:hos}. An example is given in
Fig.~\ref{fig:causality}.  Note that the simplification algorithm
of~\cite{ballester.07.jmiv} is not hierarchical either.

Instead of fixing the parameter $\lambda_s$ in the energy functional
(\eg, $\lambda_s$ in Eq~(\ref{eq:simplemumford})), we propose to
compute an individual $\lambda_s$ for each shape of the tree following
the same principle of the energy minimization. For a given
$\lambda_s$, the removability of a shape $\tau$ is based on the sign
of energy variation $\Delta E_{\lambda_s}^\tau$ in
Eq.~\eqref{eq:variation}, which is a linear decreasing function
w.r.t. $\lambda_s$ (\eg, the Eq~(\ref{eq:variationmumford}) for the
piecewise-constant Mumford-Shah functional). When $\lambda_s$ is
bigger than some value $\lambda_{s_{min}}$, $\Delta
E_{\lambda_s}^\tau$ will be negative, which implies the removal of
this shape decreases the energy functional. Thus $\lambda_{s_{min}}$
is a value of the transition for the removal decision of the
underlying shape.  Let us denote this value of transition as the
attribute function $\Attribute_{\lambda_s}$, which is given by:
\begin{equation}
 \Attribute_{\lambda_s}(\tau) = \frac{D(R'_{\tau_p}) - D(R_{\tau}) -
 D(R_{\tau_p})}{C(R_{\tau}) + C(R_{\tau_p}) -
 C(R'_{\tau_p})},
  \label{eq:lambda}
\end{equation}
For the piecewise-constant energy functional in
Eq.~\eqref{eq:simplemumford}, it is given by:
\begin{equation}
  \Attribute_{\lambda_s}(\tau) = \big( \frac{S^2(f, R_\tau)}{|R_\tau|} +
  \frac{S^2(f, R_{\tau_p})}{|R_{\tau_p}|} -\frac{S^2(f,
    R'_{\tau_p})}{|R'_{\tau_p}|} \big) \, / \, |\partial \tau|.
  \label{eq:lambdamumford}
\end{equation}

Note that for a given shape $\tau$, the attribute function
$\Attribute_{\lambda_s}(\tau)$ defined in Eq~(\ref{eq:lambda}) depends
on $R_\tau$ and $R_{\tau_p}$, which means
$\Attribute_{\lambda_s}(\tau)$ is decided by the shape $\tau$ itself,
its parent, its siblings, and its children.  Because the attribute
function $\Attribute_{\lambda_s}$ is computed under the hypothesis
that the shape $\tau$ under scrutiny is suppressed, we also need to
update $R_{\tau_p}$, and update the parenthood relationship for its
children to its parent.  These update operations will also affect the
computation of $\Attribute_{\lambda_s}$ for the parent, children and
siblings of $\tau$. So the computation order is again important. We
follow the same principle as described in
Section~\ref{subsec:llselection} to compute $\Attribute_{\lambda_s}$,
which is detailed as below:

\smallskip
\noindent
\textit{step 1:} Compute $\Attribute_{\lambda_s}$ for each shape $\tau
\in \Tree$ supposing that only the shape under scrutiny is removed,
and sort the set of shapes $\{\tau \, | \, \tau \in \Tree\}$ in
increasing order of shape meaningfulness indicated by an attribute
$\Attribute$ (\eg, $\Attribute_\nabla$).

\smallskip
\noindent
\textit{step 2:} Propagate the sorted shapes in increasing order, and
remove the shape one by one. Compute the new value
$\Attribute_{\lambda_s}$ for the underlying shape $\tau$, and update
it if the value is greater than the older one.  Update also the
parenthood relationship and the corresponding information for
$R_{\tau_p}$.

\smallskip

This attribute function $\Attribute_{\lambda_s}$ is related to the
minimization of the energy functional. It measures the persistence of
a shape to be removed under the minimization problem of
Eq~(\ref{eq:solution}). A bigger $\Attribute_{\lambda_s}(\tau)$ means
that it is more difficult to remove the shape $\tau$. Thus the
attribute function $\Attribute_{\lambda_s}$ is also some kind of
quantitative meaningfulness deduced from the energy minimization.

We use the inverse of the attribute $\Attribute_{\lambda_s}$ described
above as the final attribute function:
$\Attribute_{\lambda_s}^\downarrow(\tau) \, = \, \underset{\tau' \in
  \Tree}{\textrm{max}}\big(\Attribute_{\lambda_s}(\tau')\big) -
\Attribute_{\lambda_s}(\tau)$. The local minima of the shape space
weighted by this attribute function correspond to a set of candidate
salient level lines.  We make use of the scheme of hierarchy
transformation described in Section~\ref{subsec:ssf} to compute a
saliency map $\mathcal{M}_{\mathcal{E}}$.  This saliency map
$\mathcal{M}_{\mathcal{E}}$ represents hierarchical result of level
line selections. Each thresholding of this map
$\mathcal{M}_{\mathcal{E}}$ selects salient (of certain degree) level
lines from which a simplified image can be reconstructed.

\begin{figure}[h]
  \centering
  \begin{subfigure}[b]{0.45\linewidth}
    \centering
    \includegraphics[width=0.8\linewidth]{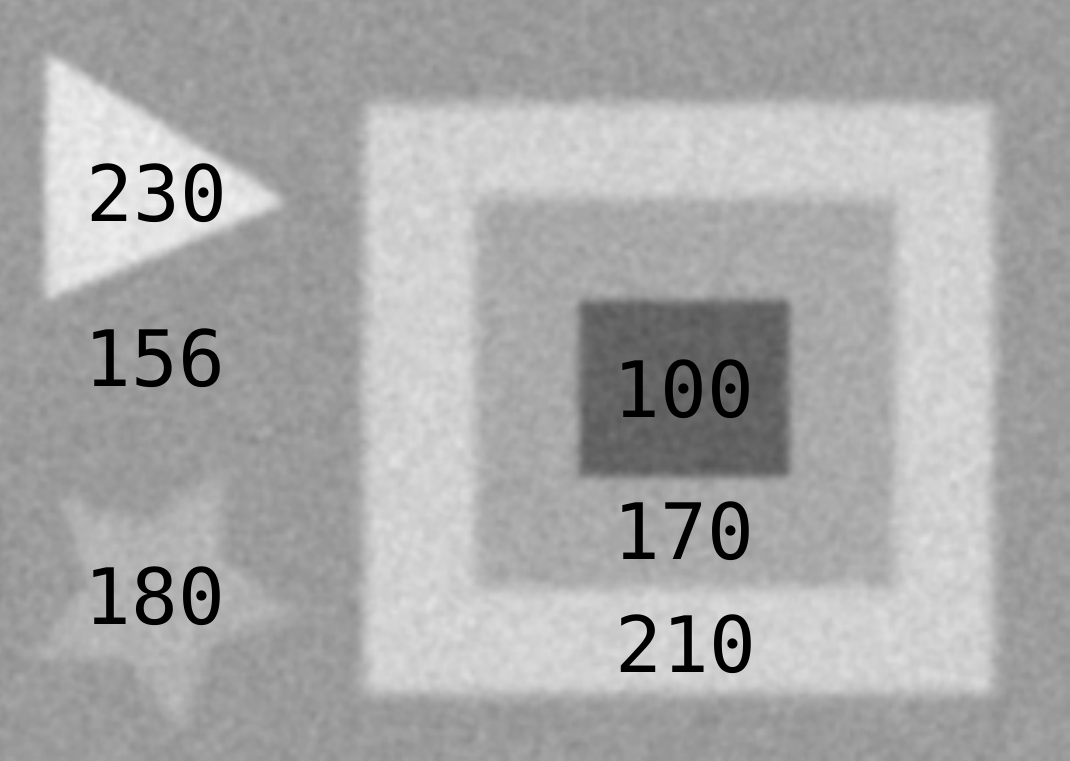}
    \caption{Input image.}
  \end{subfigure}
  ~
  \begin{subfigure}[b]{0.45\linewidth}
    \centering
    \includegraphics[width=0.9\linewidth]{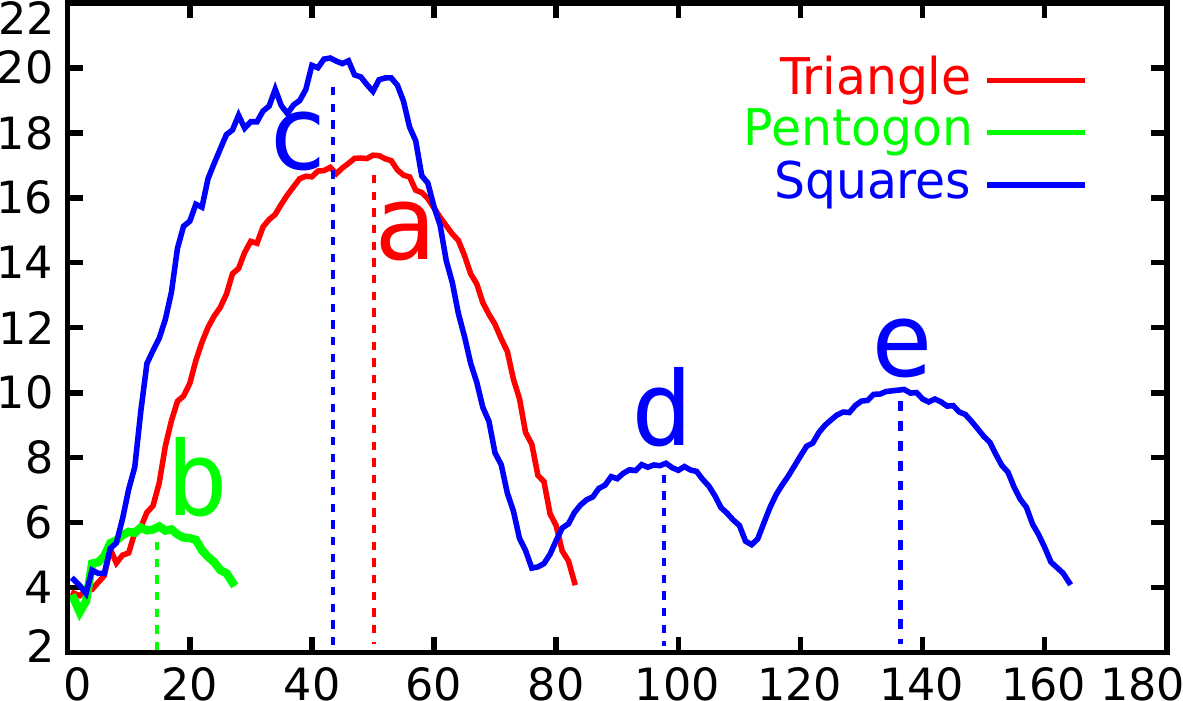}
    \caption{Gradient's magnitude $\Attribute_\nabla$.}
  \end{subfigure}

  \begin{subfigure}[b]{0.45\linewidth}
    \centering
    \includegraphics[width=0.9\linewidth]{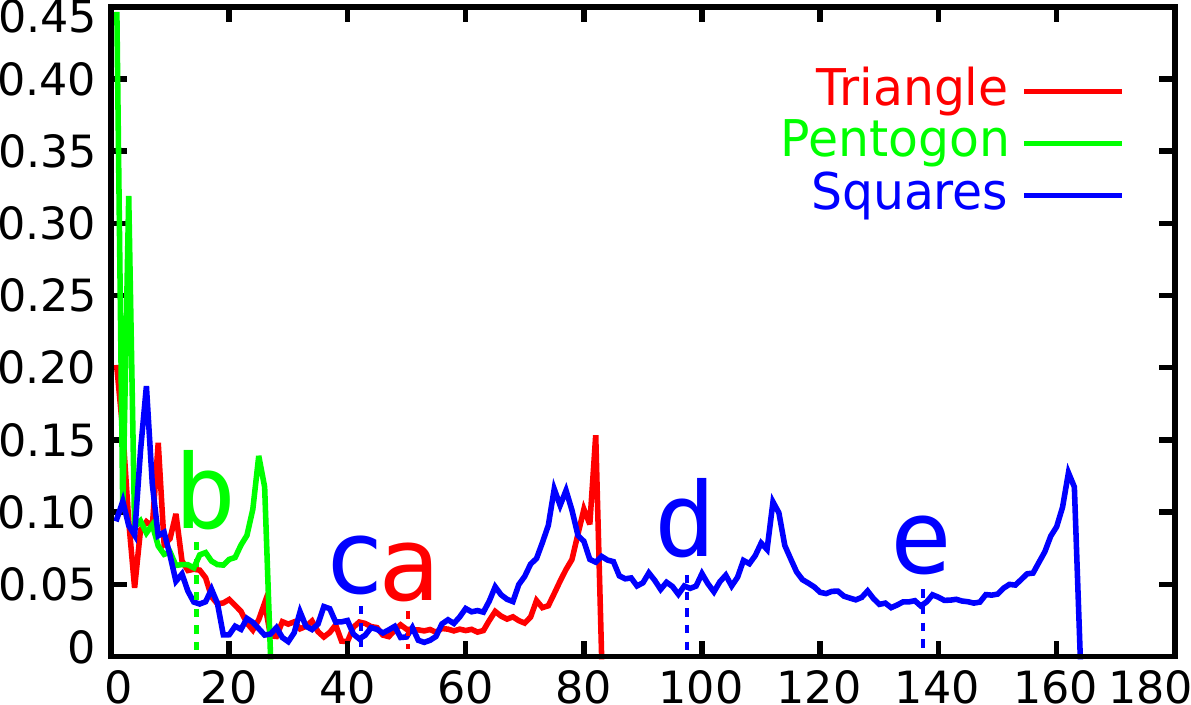}
    \caption{Initial $\Attribute_{\lambda_s}$ of step 1.}
  \end{subfigure}
  ~
  \begin{subfigure}[b]{0.45\linewidth}
    \centering
    \includegraphics[width=0.9\linewidth]{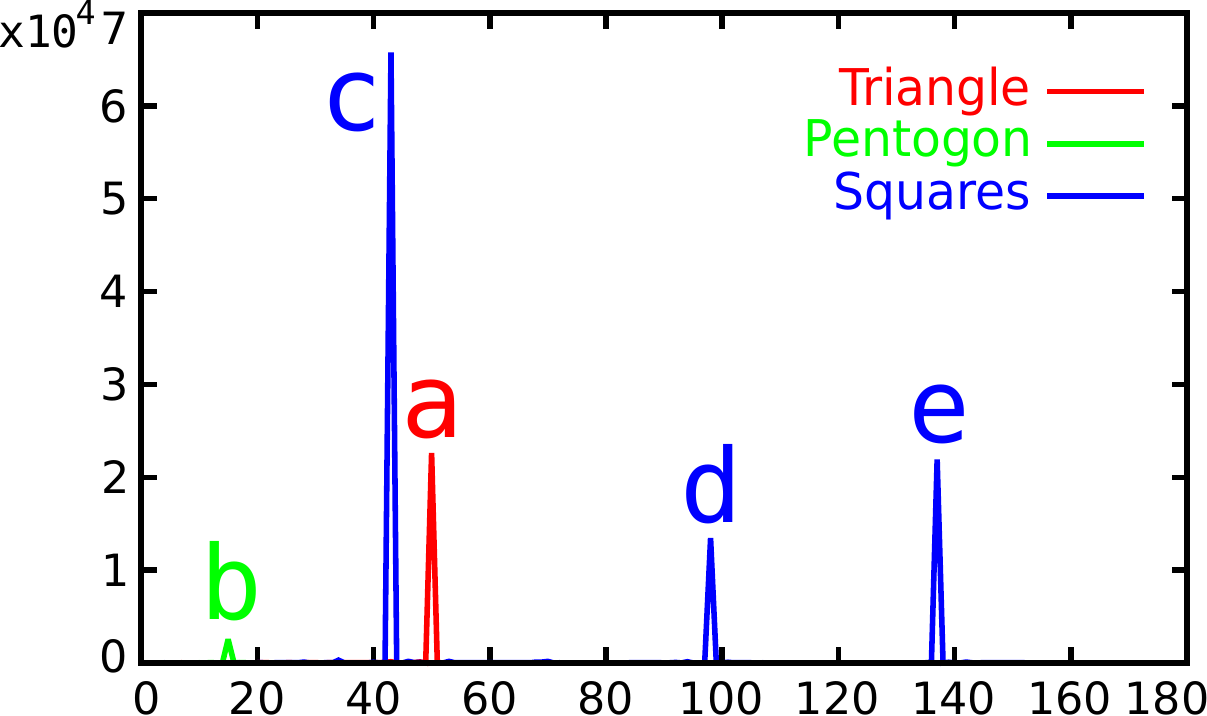}
    \caption{Final $\Attribute_{\lambda_s}$ of step 2.}
  \end{subfigure}

  \begin{subfigure}[b]{0.45\linewidth}
    \centering
    \includegraphics[width=0.8\linewidth]{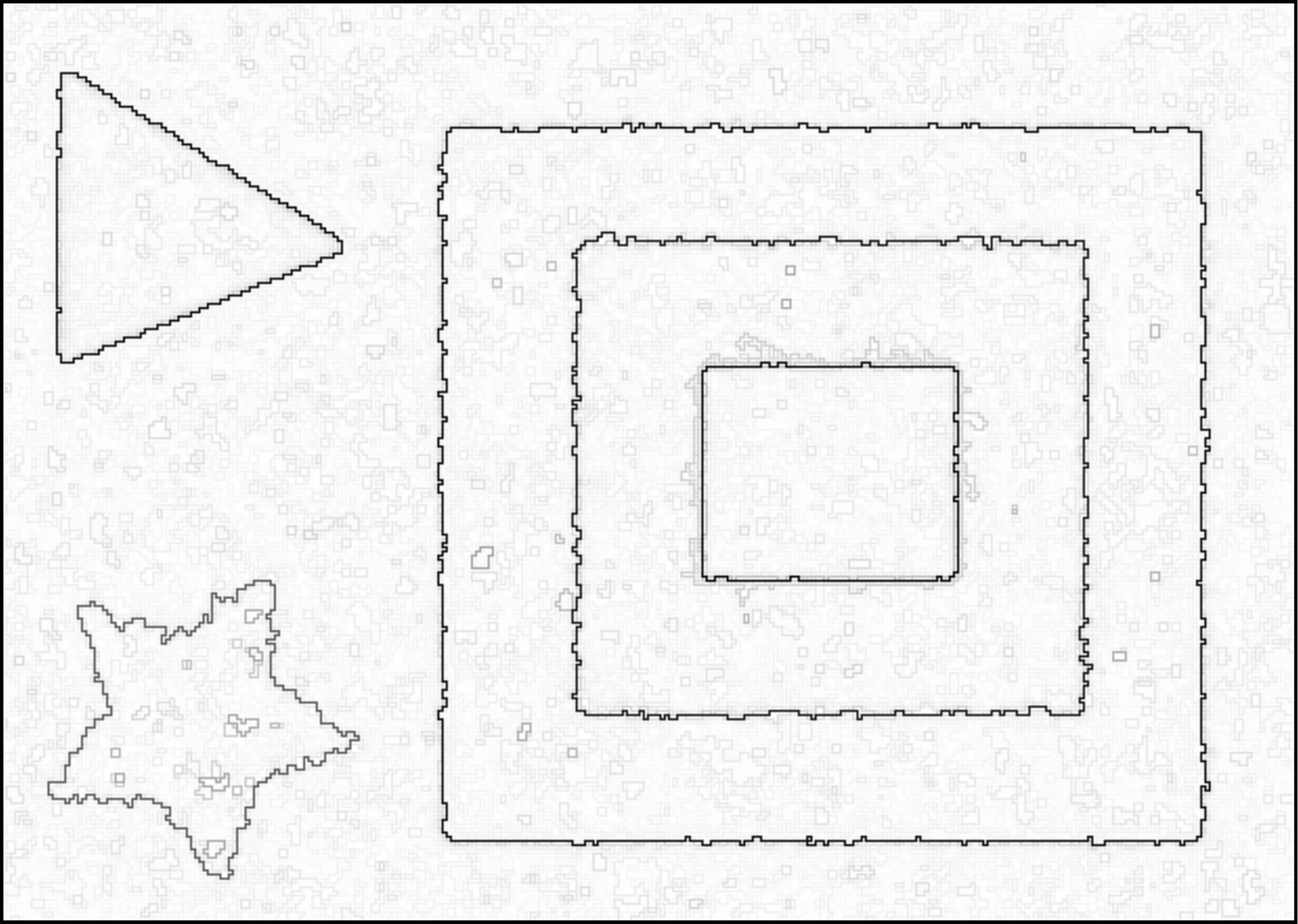}
    \caption{Saliency map $\map_\extinct$.}
  \end{subfigure}
  ~
  \begin{subfigure}[b]{0.45\linewidth}
    \centering
    \includegraphics[width=0.8\linewidth]{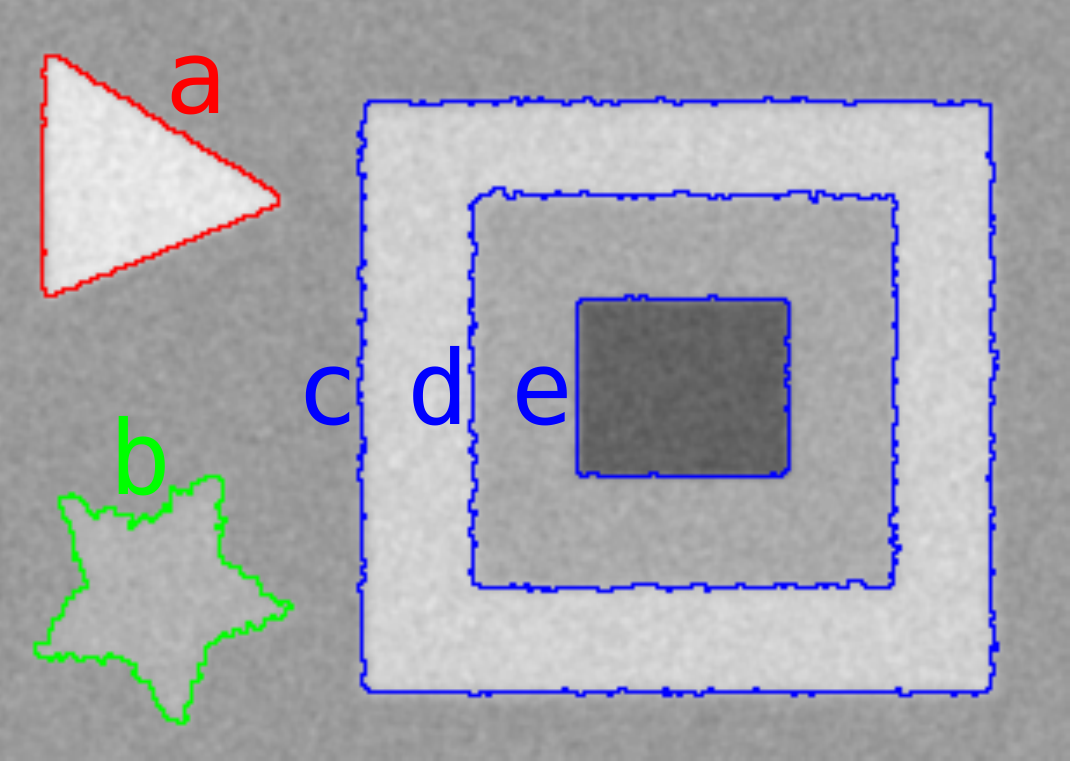}
    \caption{A segmentation.}
  \end{subfigure}
  \caption{An example of the proposed scheme on a synthetic
    image. (b-d): Evolution of attribute value starting from leaf
    regions (left end of each curve) inside the triangle (red),
    pentagon (green), and squares (blue) to the root of the tree
    (right end of each curve). Note that the length of these three
    branches is different (it explains that the root node appears at
    different abscissas.)}
  \label{fig:example}
\end{figure}

An example of the proposed scheme on a synthetic image is illustrated
in Fig.~\ref{fig:example}. The input image in
Fig.~\ref{fig:example}~(a) is both blurred and noisy. This blurring is
also visible in Fig.~\ref{fig:example}~(b) that illustrates the
evolution of the average of gradient's magnitude $\Attribute_\nabla$
along the contours of shapes starting from regions inside the
triangle, pentagon, and square regions to the root of the tree. The
evolution of the initial values of the Attribute
$\Attribute_{\lambda_s}$ obtained at step 1 on the same branches of
the tree are provided in Fig.~\ref{fig:example}~(c). It is not
surprising that those initial values $\Attribute_{\lambda_s}$ are not
effective to measure the {\bluetext importance} of the shapes: indeed,
this is due to the very small size of each region in
$\{R_\Tree\}$. The evolution of the final values of the attribute
$\Attribute_{\lambda_s}$ (of step 2) is depicted in
Fig.~\ref{fig:example}~(d). We can see that the significant regions
are highlighted by $\Attribute_{\lambda_s}$. This experiment also
demonstrates the relevance of the increasing order of average of
gradient's magnitude along the contour $\Attribute_\nabla$ as a
criterion to update the value of $\Attribute_{\lambda_s}$. The
saliency map $\map_\extinct$ using the attribute
$\Attribute_{\lambda_s}$ and one of the possible segmentations that
can be obtaiend by thresholding $\map_\extinct$ are depicted in
Fig.~\ref{fig:example}.~(e) and~(f).

\subsection{Implementation}
\label{subsec:algorithm}

The proposed method is composed of {\bluetext three} main steps: 1)
Construction of the tree of shapes and computation of the attribute function
$\Attribute_{\lambda_s}$; 2) Computation of the extinction values
$\extinct$; 3) Computation of the saliency map $\map_\extinct$.


Once we have the tree structure $\Tree$ represented by the parenthood
image $parent$, and the corresponding information $\Attribute$ for
each node of the tree, we are able to compute the attribute function
$\Attribute_{\lambda_s}$.  {\bluetext Note that the information
  $\Attribute$ can be the area $A$, the sum of gray levels $S_f$, the
  contour length $L$, or the sum of gradient's magnitude along the
  contour $S_\nabla$.}  The computation of this attribute function
$\Attribute_{\lambda_s}$ is performed while computing the tree, and is
detailed in Algorithm~\ref{algo:attribute}. We start by computing the
initial values of attribute $\Attribute_{\lambda_S}$ according to
Eq~\eqref{eq:lambdamumford} by considering that only the underlying
shape is removed (see line 11). Then we sort the shapes by increasing
order of the average of gradient's magnitude along the shape contour
$\Attribute_\nabla$. We process the shapes in this order. For each
underlying shape $\tau$, we compute a new value according to
Eq~\eqref{eq:lambdamumford} (see line 15), and update the value
$\Attribute_{\lambda_s}(\tau)$ if the new value is greater. Then we
remove the shape $\tau$ from the tree and update the tree structure as
well as the corresponding information.

\begin{algorithm}[t]
  COMPUTE\_ATTRIBUTE($parent$, $\Tree$, $\Attribute$) \\
  \ForAll{$\tau \in \Tree$}{$A_R(\tau) \leftarrow A(\tau)$, $S_{f, R}(\tau) \leftarrow S_f(\tau)$; \\
    \lIf{$\tau \neq parent(\tau)$}{$Ch(parent(\tau)).insert(\tau)$};}
  \ForAll{$\tau \in \Tree$}{\If{$\tau \neq parent(\tau)$}{$A_R\big(parent(\tau)\big) \leftarrow A_R(parent\big(\tau)\big) - A(\tau)$; \\ $S_{f,R}\big(parent(\tau)\big) \leftarrow S_{f,R}(parent\big(\tau)\big) - S_f(\tau)$; 
  }}
  \ForAll{$\tau \in \Tree$}{
   $\Attribute_\nabla(\tau) \leftarrow S_\nabla(\tau)/L(\tau)$, $\tau_p \leftarrow parent(\tau)$; \\
    $\Attribute_{\lambda_s}(\tau) \leftarrow \big(\frac{S_{f, R}^2(\tau)}{A_R(\tau)} +  \frac{S_{f, R}^2(\tau_p)}{A_R(\tau_p)} - \frac{(S_{f, R}(\tau_p) + S_{f, R}(\tau))^2}{A_R(\tau_p) + A_R(\tau)}\big)/L(\tau)$;
  }
  $\mathcal{R}_\Tree \leftarrow $ SORT\_NODES($\Tree$, $\Attribute_\nabla$)\\

  \For{$i \leftarrow 0$ \KwTo $N_\Tree$}{$\tau \leftarrow R_\Tree(i)$, $\tau_p \leftarrow parent(\tau)$; \\
    $\lambda_t \leftarrow \big(\frac{S_{f, R}^2(\tau)}{A_R(\tau)} +  \frac{S_{f, R}^2(\tau_p)}{A_R(\tau_p)} - \frac{(S_{f, R}(\tau_p) + S_{f, R}(\tau))^2}{A_R(\tau_p) + A_R(\tau)}\big)/L(\tau)$; \\
  \lIf{$\lambda_t > \Attribute_{\lambda_s}(\tau)$}{$\Attribute_{\lambda_s}(\tau) \leftarrow \lambda_t$;} \\
  $Ch(\tau_p).remove(\tau)$; \\
  \ForAll{$\tau_c \in Ch(\tau)$}{$parent(\tau_c) \leftarrow \tau_p$; \\
    $Ch(\tau_p).insert(\tau_c)$;}
  $A_R(\tau_p) \leftarrow A_R(\tau_p) + A_R(\tau)$; \\
  $S_{f,R}(\tau_p) \leftarrow S_{f,R}(\tau_p) + S_{f,R}(\tau)$; \\
  }

  \Return{$\Attribute_{\lambda_s}$}
  \caption{Computation of attribute function
    $\Attribute_{\lambda_s}$. During the tree computation, we also
    compute four attribute information $\Attribute$: region size $A$,
    region's contour length $L$, sum of gray level $S_f$ inside the
    region, and sum of gradient's magnitude along the region contour
    $S_\nabla$.}
  \label{algo:attribute}
\end{algorithm}

The {\bluetext second} step is to compute the extinction values
$\extinct$ for the local minima of the tree-based shape space weighted
by the attribute $\Attribute_{\lambda_s}^\downarrow$. This is achieved
thanks to a min-tree representation $\Tree\Tree$ constructed on the
tree-based shape space. The algorithm is described in
Algorithm~\ref{algo:extinction}. The image $original\_min$ tracks the
smallest local minimum inside a lower level connected component $CC$
of the tree-based shape space. For each local minimum shape $\tau$,
the lowest $CC$ that contains $\tau$ and a smaller minimum is the
lowest ancestor node whose smallest local minimum shape is different
from $\tau$.

\begin{algorithm}[t]
  COMPUTE\_EXTINCTION\_VALUE($\Tree$, $\Attribute$) \\
  $(parent_\Tree, \mathcal{R}_\Tree) \leftarrow$ COMPUTE\_TREE($\Attribute$); \\
  \lForAll{$\tau \in \Tree$}{$original\_min(\tau) \leftarrow$ {\em undef};}\\
  \For{$i \leftarrow 0$ \KwTo $N_\Tree$}{$\tau \leftarrow R_\Tree(i)$, $\tau_p \leftarrow parent_\Tree(\tau)$; \\
    \lIf{$original\_min(\tau) =$ {\em undef}}{$original\_min(\tau) = \tau$;} \\
    \uIf{$original\_min(\tau_p) =$ {\em undef}}{$original\_min(\tau_p) \leftarrow original\_min(\tau)$;}
    \Else{
      \If{$\Attribute(original\_min(\tau_p)) > \Attribute(original\_min(\tau))$}{$original\_min(\tau_p) \leftarrow original\_min(\tau)$;}
    }
   }
  \ForAll{$\tau \in \Tree$}{
    \uIf{$\tau$ is not a local minimum}{$\extinct(\tau) \leftarrow 0$;}
    \Else{
      $\tau_p \leftarrow parent_\Tree(\tau)$; \\
      \While{$original\_min(\tau_p) = \tau \And \tau_p \neq parent_\Tree(\tau_p)$}{$\tau_p \leftarrow(parent_\Tree(\tau_p))$;}
      $\extinct(\tau) \leftarrow \Attribute(\tau_p) - \Attribute(\tau)$;
    }
  }
  \Return{$\extinct$}
  \caption{Computation of extinction values $\extinct$ on a
    tree-based shape space weighted by an attribute $\Attribute$. The
    image $original\_min$ tracks the smallest local minimum shape
    inside a connected component of $\Tree\Tree$. Note that
    $parent_\Tree$ encodes the min-tree $\Tree\Tree$ constructed on
    the shape space.}
  \label{algo:extinction}
\end{algorithm}

\begin{figure}
  \centering
  \includegraphics[width=0.25\linewidth]{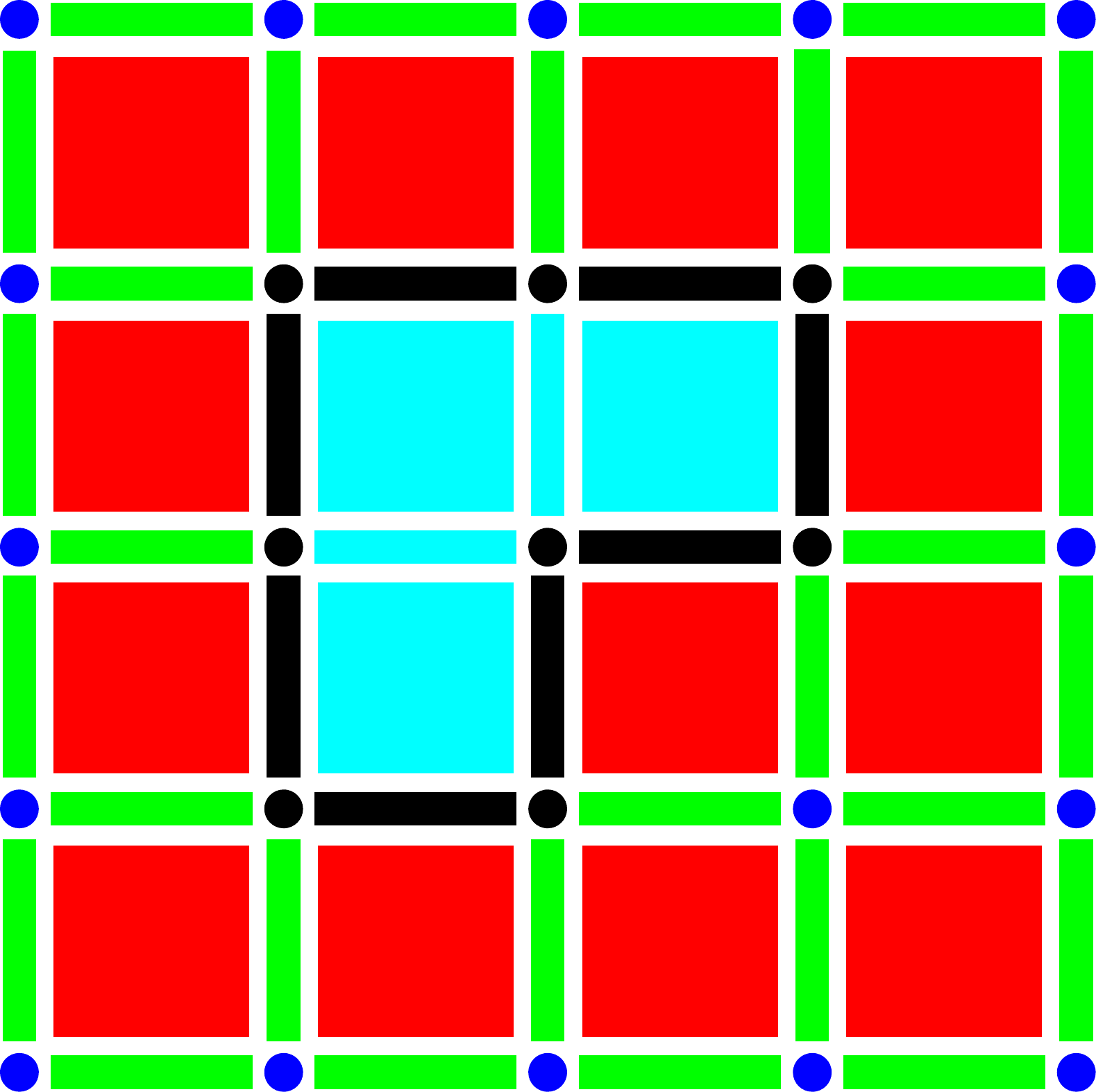}
  \caption{Materialization of pixels with 0-faces (blue
    disks), 1-faces (green strips), and 2-faces (red squares). The
    original pixels are the 2-faces, the boundaries are materialized
    with 0-faces and 1-faces. The contour of the cyan region is
    composed of black 1-faces and 0-faces.}
  \label{fig:interpolation}
\end{figure}

To compute the final saliency map, we rely on the Khalimsky's grid,
{\bluetext proposed by~\cite{khalimsky.90.tia}, and depicted in
  Fig.~\ref{fig:interpolation}}; it is composed of $0$-faces (points),
$1$-faces (edges) and $2$-faces (pixels).  The saliency map
$\map_{\extinct}$ is based on the extinction values, where we weigh
the boundaries ($0$-faces and $1$-faces) of each shape by the
corresponding extinction value $\extinct$. More precisely, we weigh
each $1$-face $e$ (\resp $0$-face $o$) by the maximal extinction value
of the shapes whose boundaries contain $e$ (\resp $o$). The algorithm
is given in Algorithm~\ref{algo:saliency}. It relies on two images
$\mathit{appear}$ and $\mathit{vanish}$ defined on the 1-faces that
are computed during the tree construction. The value
$\mathit{appear}(e)$ encodes the smallest region $\mathcal{N}_a$ in
the tree whose boundary contains the 1-face $e$, while
$\mathit{vanish}(e)$ denotes the smallest region $\mathcal{N}_v$ that
contains the 1-face $e$ inside it.

\begin{algorithm}[t]
  COMPUTE\_SALIENCY\_MAP($f$) \\
  $(parent, \Tree, \Attribute) \leftarrow$ COMPUTE\_TREE(f); \\
  $\Attribute_{\lambda_s} \leftarrow$ COMPUTE\_ATTRIBUTE($parent$, $\Tree$, $\Attribute$); \\
  $\lambda_s^M \leftarrow 0$; \\
  \lForAll{$\tau \in \Tree$}{$\lambda_s^M \leftarrow \max\big(\lambda_s^M, \Attribute_{\lambda_s}(\tau)\big)$;} \\
  \lForAll{$\tau \in \Tree$}{$\Attribute_{\lambda_s}^\downarrow \leftarrow \lambda_s^M - \Attribute_{\lambda_s}(\tau)$;} \\
  $\extinct \leftarrow$ COMPUTE\_EXTINCTION\_VALUE($\Tree, \Attribute_{\lambda_s}^\downarrow$); \\
  \lForAll{$1$-face $e$}{$\mathcal{M}_\extinct(e) \leftarrow 0$;} \\
  \ForAll{$e$}{
    $N_a \leftarrow appear(e), N_v \leftarrow vanish(e)$;\\
    \While{$N_a \neq N_v$}{
      $\mathcal{M}_\extinct(e) \leftarrow \textbf{max}\big(\extinct(N_a), \mathcal{M}_\extinct(e)\big)$,
        $N_a \leftarrow parent(N_a);$
      }
    }
    \ForAll{$0$-face $o$}{$\mathcal{M}_\extinct(o) \leftarrow$ \textbf{max}\big($\mathcal{M}_\extinct(e_1), \mathcal{M}_\extinct(e_2), \mathcal{M}_\extinct(e_3), \mathcal{M}_\extinct(e_4) \big)$;}
    \Return{$\mathcal{M}_\extinct$}
  \caption{Computation of saliency map $\map_\extinct$
    representing a hierarchical result of level line selections. The
    1-faces $e_1, e_2, e_3, e_4$ are the 1-faces adjacent to $o$.}
  \label{algo:saliency}
\end{algorithm}

\begin{bluetext}
  We refer the reader to~\cite{najman.06.itip, berger.07.icip,
    carlinet.14.itip, geraud.13.ismm} for details about the tree
  construction, and to~\cite{xu.15.ismm} for details about the
  efficient computation of some information $\Attribute$ (namely $A$,
  $S_f$, $L$, and $S_\nabla$).
\end{bluetext}



\section{Illustrations and experiments}
\label{sec:illustration}

In this section, we illustrate our proposed general framework with a
simple segmentation model: piecewise-constant Mumford-Shah
model. Using some more evolved energy functional will be one of our
future work. For generic natural images, contours of significant
objects usually coincide with segments of level lines. Our proposed
method yields a hierarchical simplification rather than a hierarchical
segmentation.  So only qualitative illustrations are depicted
in~Section~\ref{subsec:hierarchy} for some images taken from the
BSDS500 dataset introduced in~\cite{arbelaez.11.pami}. For the
Weizmann segmentation database proposed by~\cite{alpert.12.pami}, the
objects' contours coincide with almost full level lines. Our method
provides a hierarchical segmentation. Quantitative results using the
associated evaluation framework are depicted in
Section~\ref{subsec:segmentation}.


\subsection{Hierarchical color image pre-segmentation}
\label{subsec:hierarchy}

In Fig.~\ref{fig:simpcolor}, we test our method on color images in the
segmentation evaluation database proposed
in~\cite{alpert.12.pami}. Each image contains two objects to be
segmented. We use the color tree of shapes proposed
by~\cite{carlinet.15.itip}, where the input image $f$ in
  Eq~\eqref{eq:simplemumford} is a color image. A high parameter
value $\lambda_s = 8000$ is used, and the grain filter proposed
in~\cite{monasse.00.itip} is applied to get rid of too tiny shapes
(\eg, smaller than 10 pixels).  Less than 100 level lines are
selected, which results in a ratio of level line selection around
1157. These selected level lines form less than 200 regions in each
image. The simplified images illustrated in Fig.~\ref{fig:simpcolor}
are obtained by taking the average color inside each region, where the
boundaries between salient regions remain intact. Finding an actual
segmentation becomes a lot easier with such a pre-segmentation. The
extinction-based saliency maps $\map_\extinct$ using the attribute
$\Attribute_{\lambda_s}^\downarrow$ are depicted on the bottom of this
figure. They represent hierarchical pre-segmentations.

Some illustrations of the extinction-based saliency map
$\map_\extinct$ applied on images in the dataset of
BSDS500~\cite{arbelaez.11.pami} are also shown in
Fig.~\ref{fig:hierarchicalsimplify}. Again, the input image $f$ is a
color image, and the color tree of shapes is used. As shown in
Fig.~\ref{fig:hierarchicalsimplify}, {\bluetext salient} level lines
are highlighted in $\map_\extinct$ employing the attribute
$\Attribute_{\lambda_s}^\downarrow$.  Hierarchical image
simplification results can then be obtained by thresholding
$\map_\extinct$. The saliency maps for all the images in the dataset
of BSDS500 is available on
\url{http://publications.lrde.epita.fr/xu.hierarchymsll}.

\newcommand{\threescale}{0.31} 
\begin{figure}[t]
  \centering
  \begin{subfigure}[b]{\threescale\linewidth}
    \centering
    \includegraphics[width=1.0\linewidth]{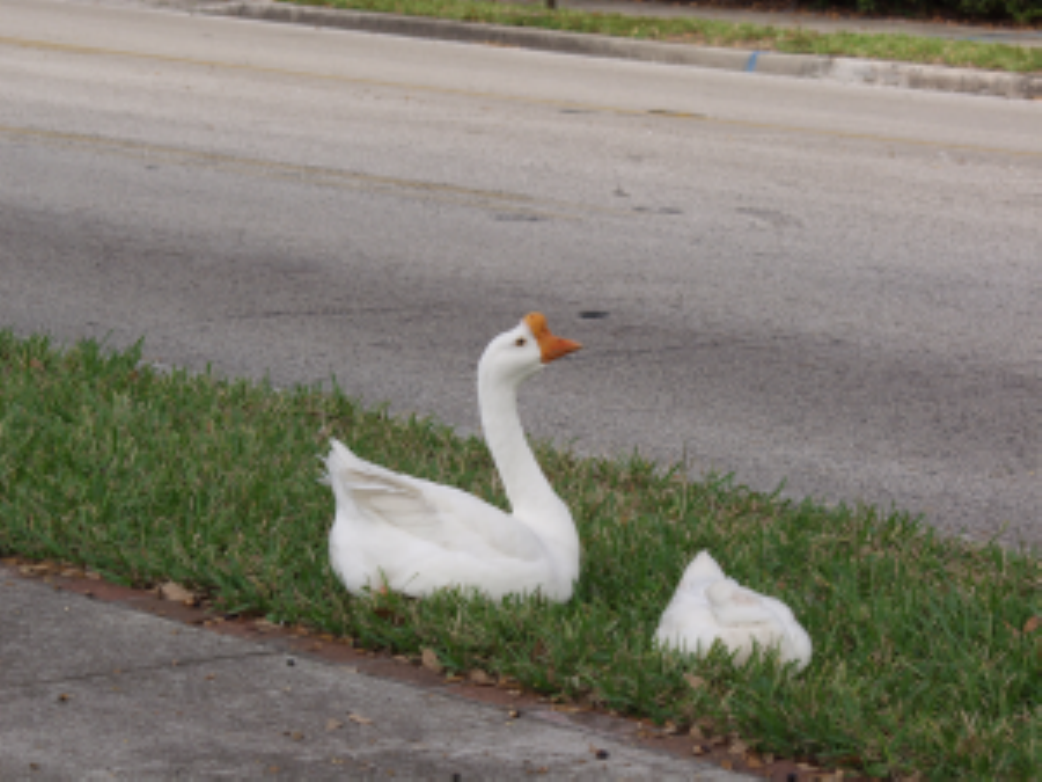}
  \end{subfigure}
  ~
  \begin{subfigure}[b]{\threescale\linewidth}
    \centering
    \includegraphics[width=1.0\linewidth]{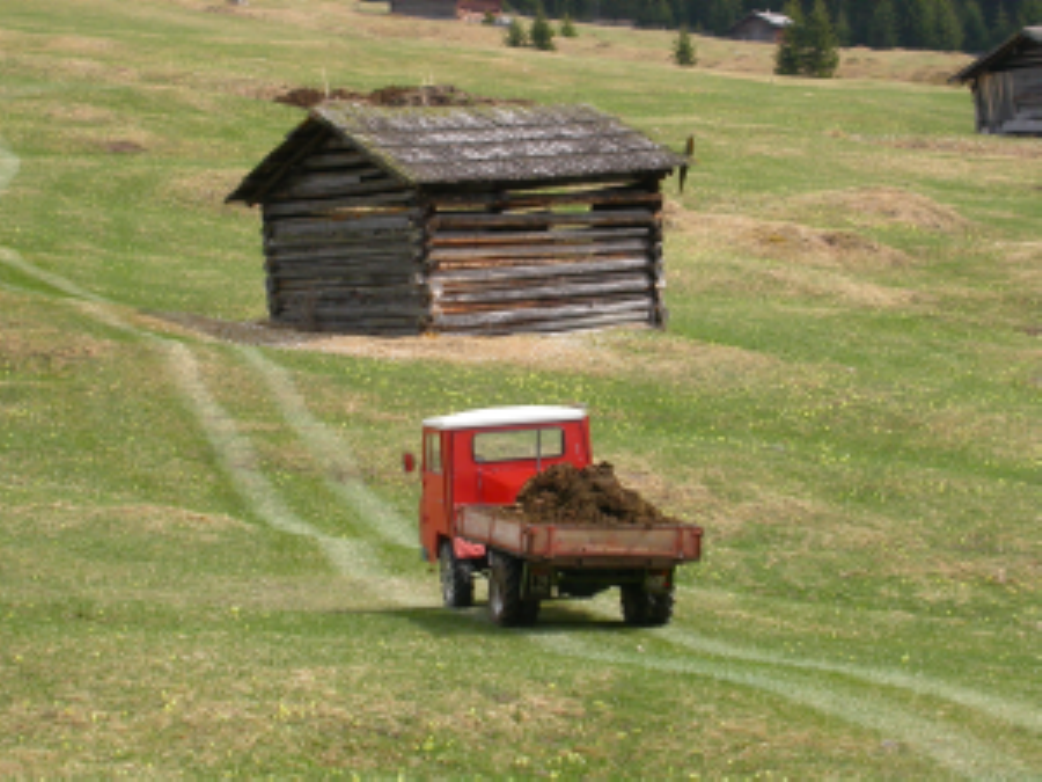}
  \end{subfigure}
  ~
  \begin{subfigure}[b]{\threescale\linewidth}
    \centering
    \includegraphics[width=1.0\linewidth]{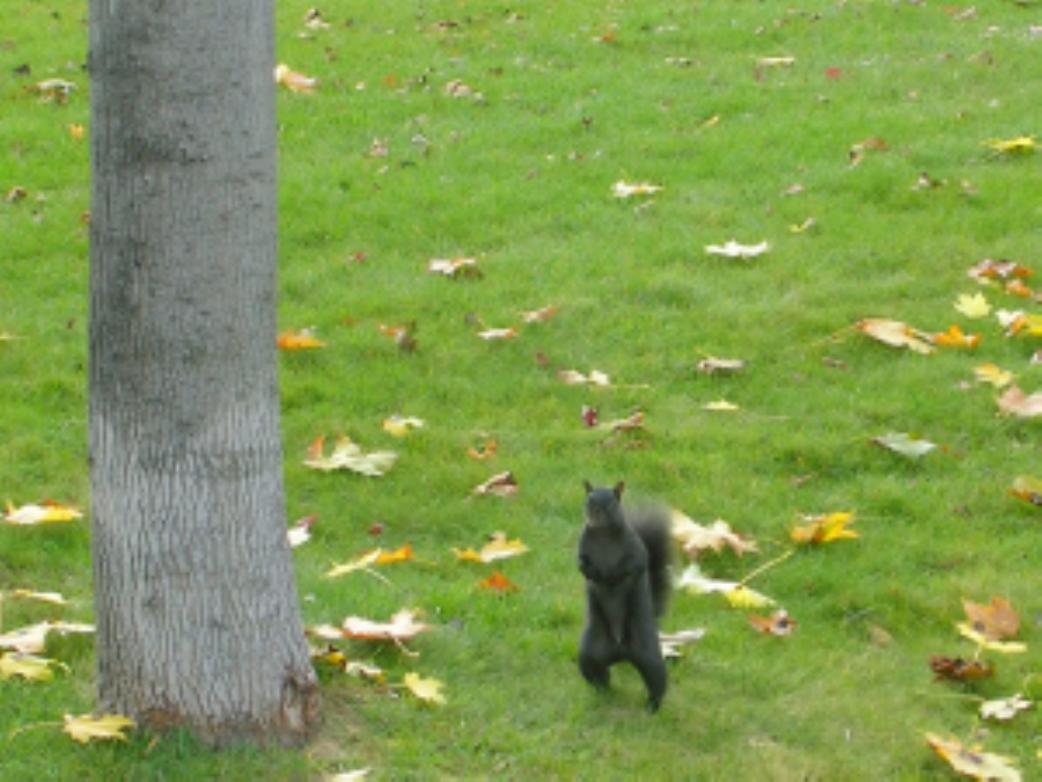}
  \end{subfigure}

  \vspace{2mm}
  \begin{subfigure}[b]{\threescale\linewidth}
    \centering
    \includegraphics[width=1.0\linewidth]{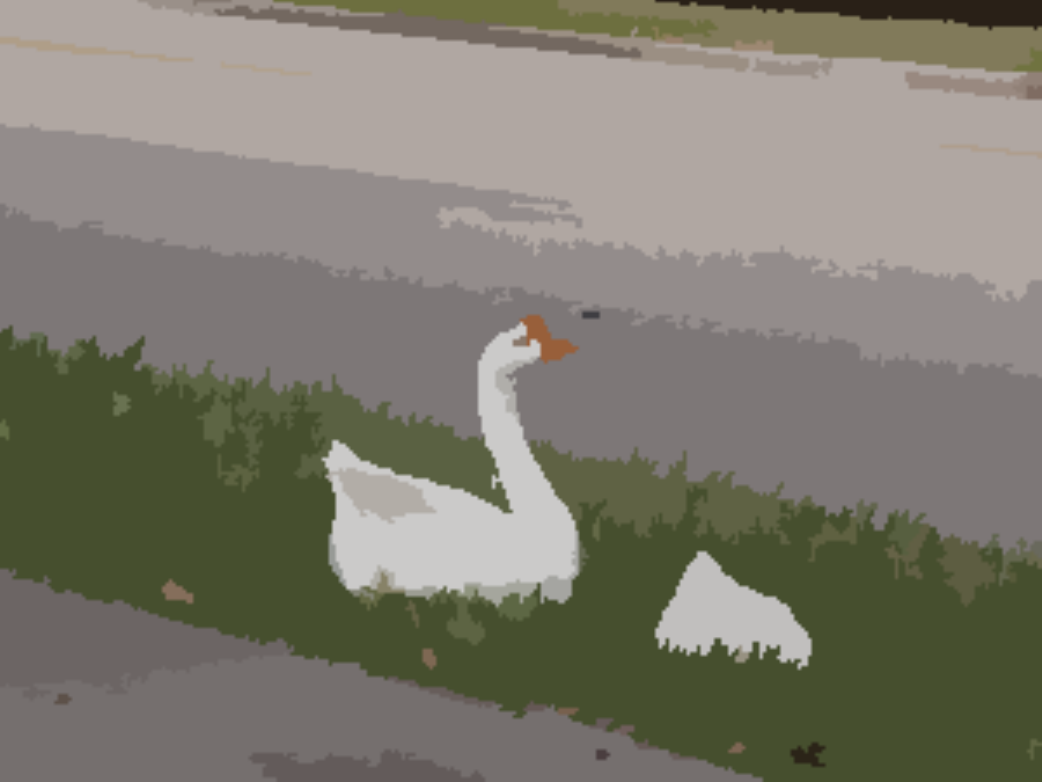}
  \end{subfigure}
  ~
  \begin{subfigure}[b]{\threescale\linewidth}
    \centering
    \includegraphics[width=1.0\linewidth]{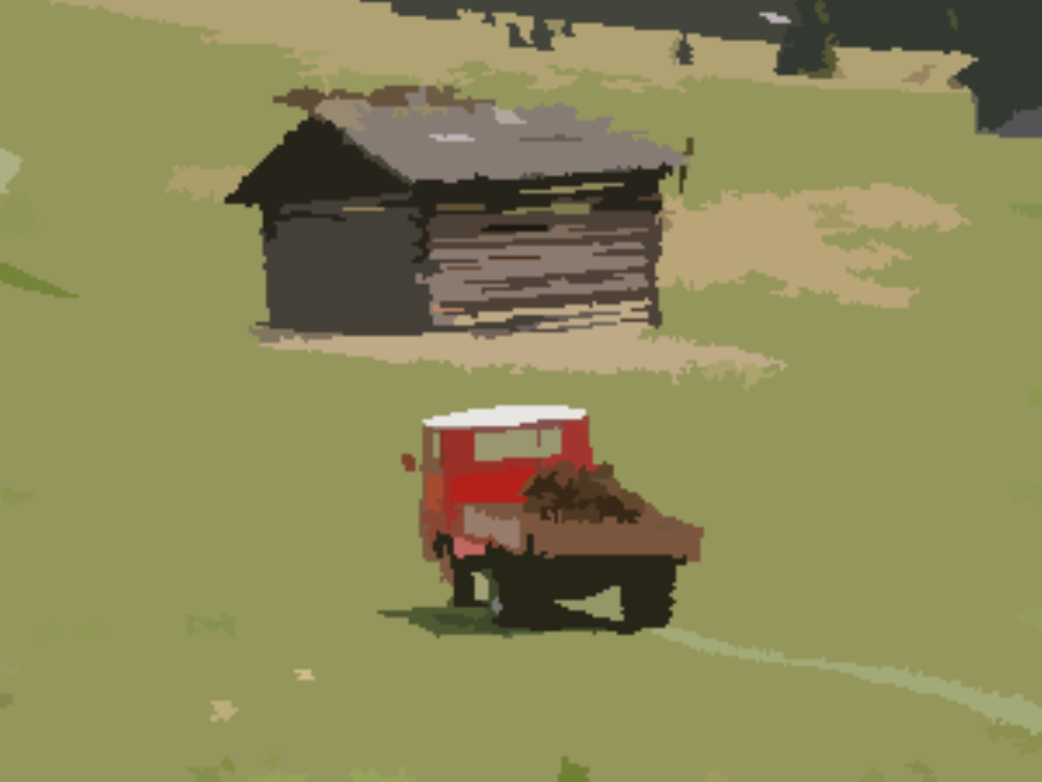}
  \end{subfigure}
  ~
  \begin{subfigure}[b]{\threescale\linewidth}
    \centering
    \includegraphics[width=1.0\linewidth]{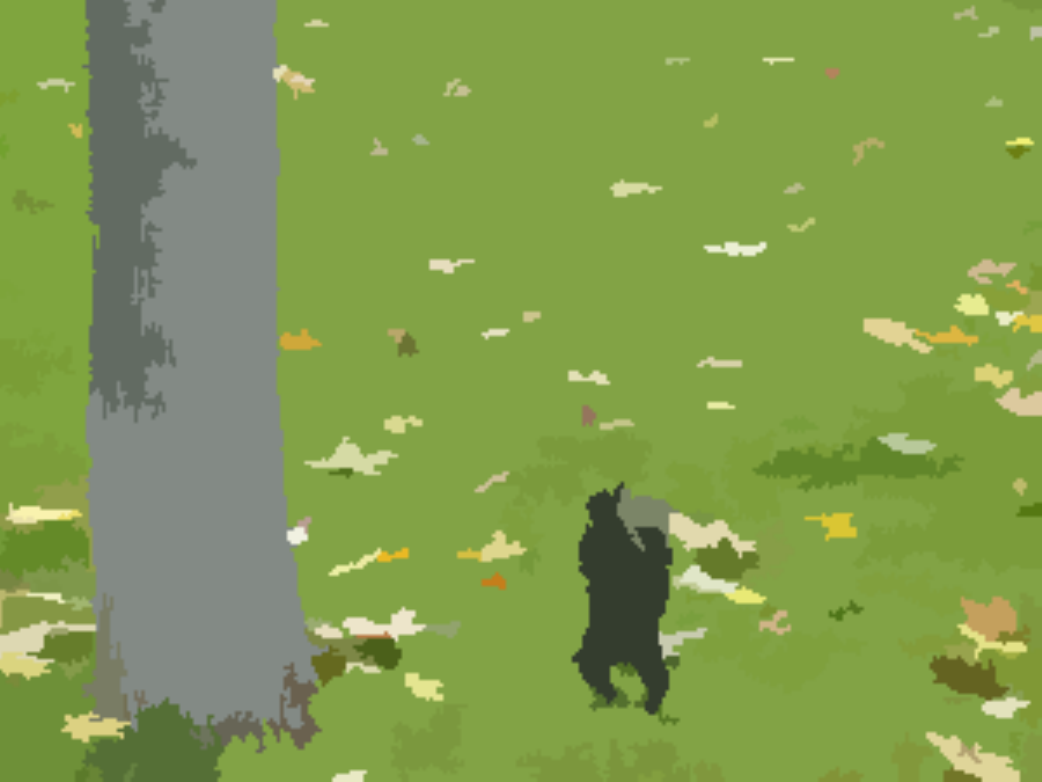}
  \end{subfigure}

    \vspace{2mm}
  \begin{subfigure}[b]{\threescale\linewidth}
    \centering
    \includegraphics[width=1.0\linewidth]{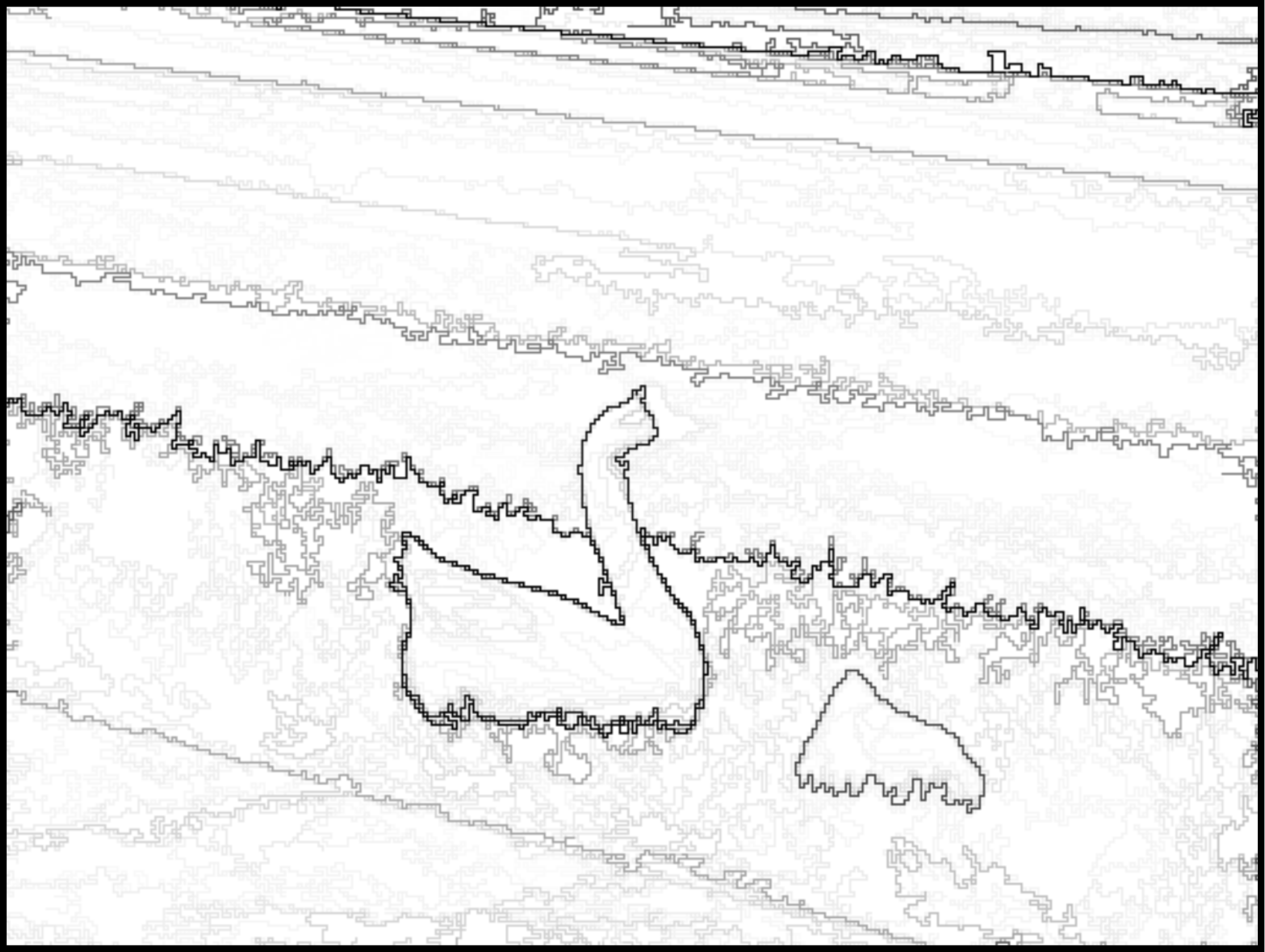}
  \end{subfigure}
  ~
  \begin{subfigure}[b]{\threescale\linewidth}
    \centering
    \includegraphics[width=1.0\linewidth]{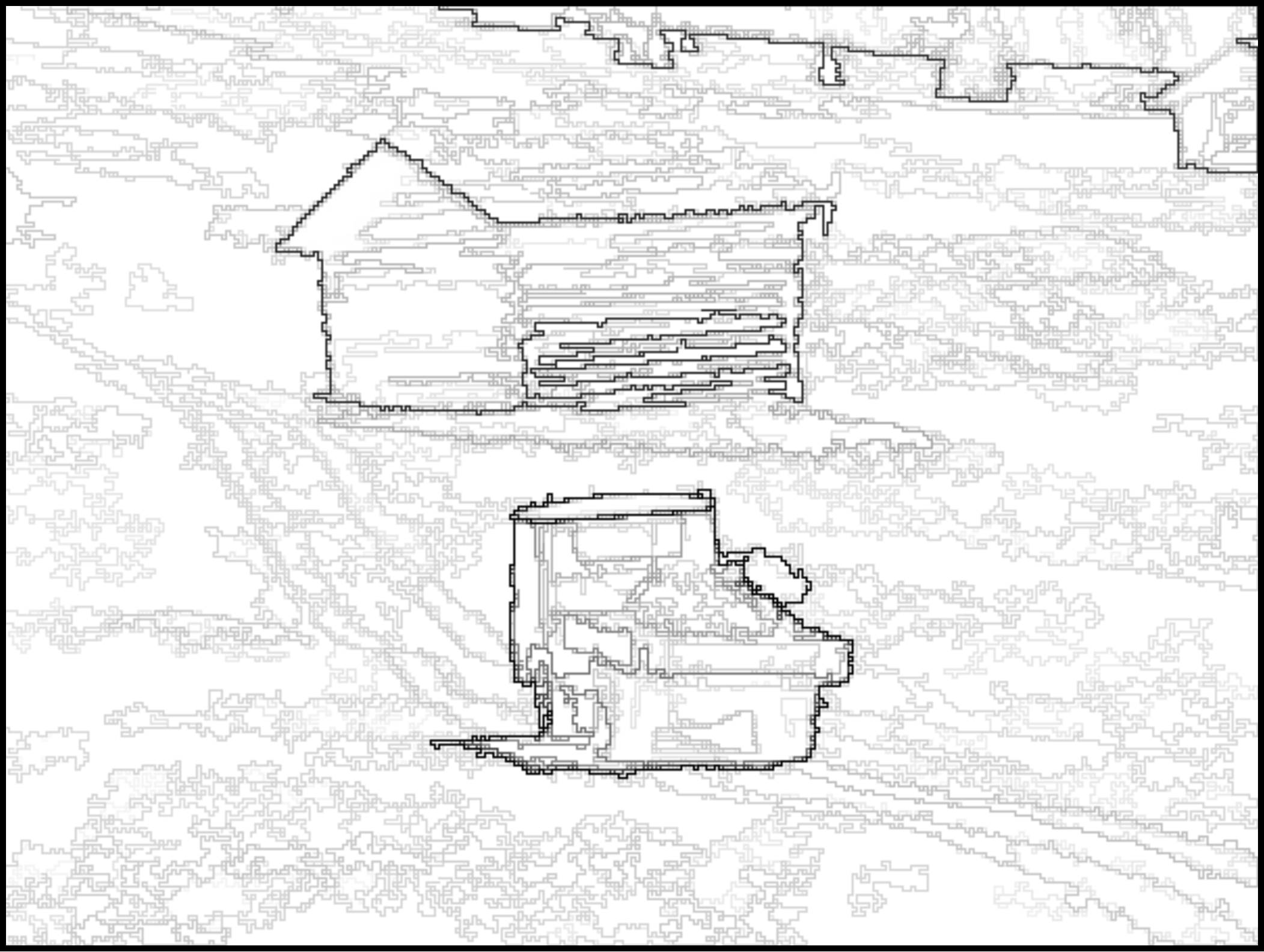}
  \end{subfigure}
  ~
  \begin{subfigure}[b]{\threescale\linewidth}
    \centering
    \includegraphics[width=1.0\linewidth]{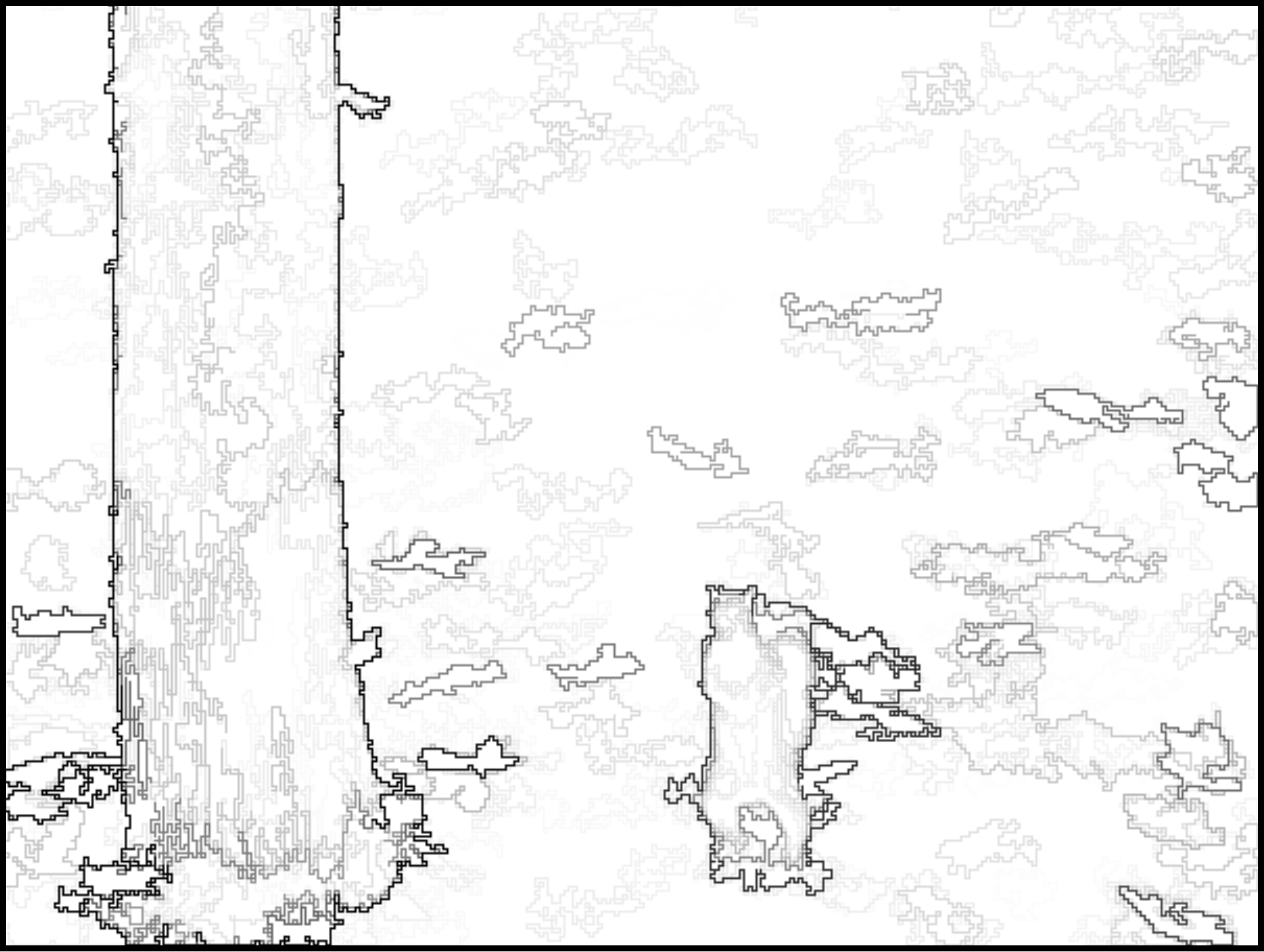}
  \end{subfigure}
  \caption{Some pre-segmentation results obtained with our proposed
    method on the segmentation evaluation database
    in~\cite{alpert.12.pami}. Top: input images; Middle:
    pre-segmentations obtained with the simplification method; Bottom:
    inverted saliency maps for hierarchical simplifications. }
  \label{fig:simpcolor}
\end{figure}

\begin{figure}[ht]
  \centering
  \begin{subfigure}{\threescale\linewidth}
    \centering
    \includegraphics[width=1.0\linewidth]{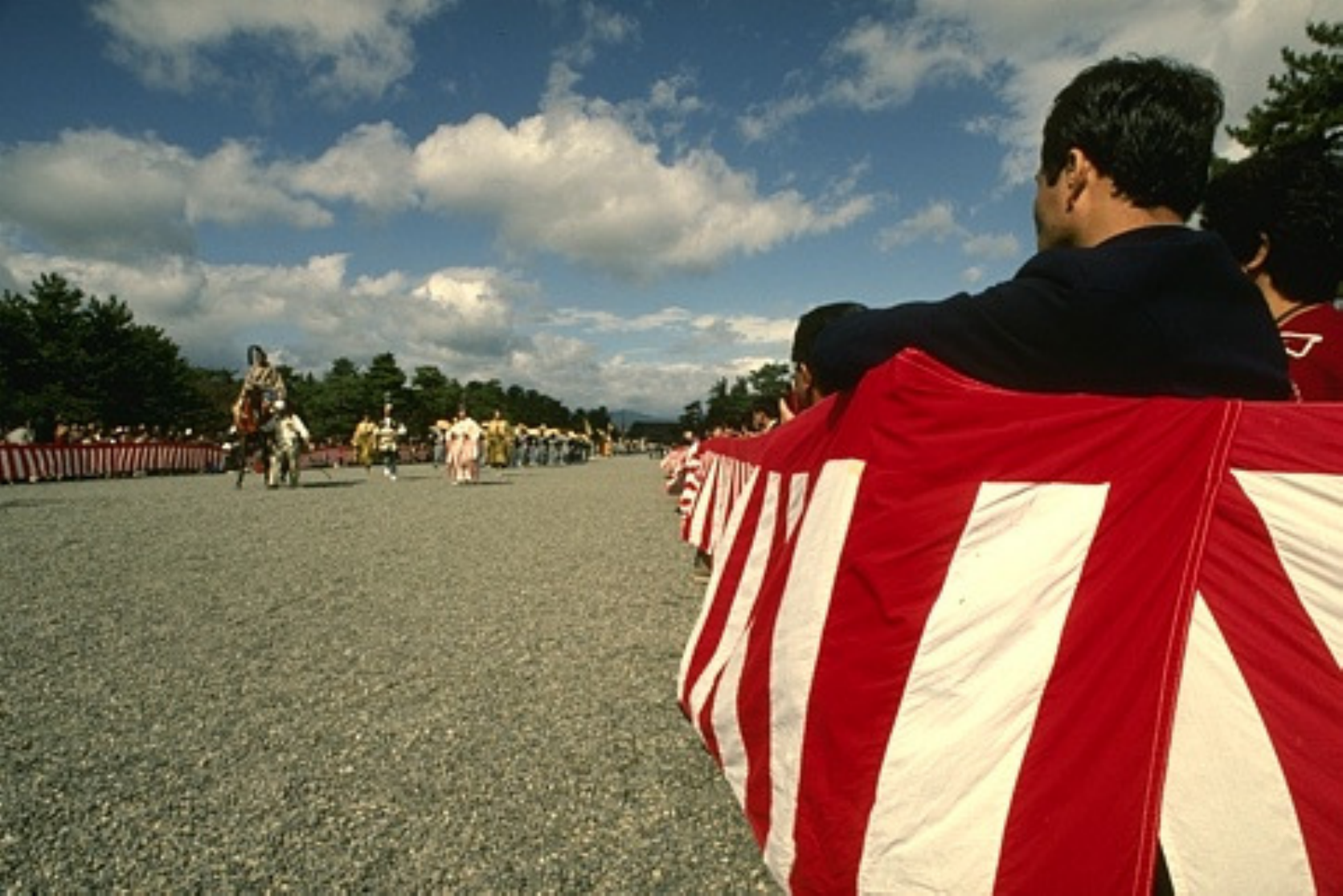}
  \end{subfigure}
  ~
  \begin{subfigure}{\threescale\linewidth}
    \centering
    \includegraphics[width=1.0\linewidth]{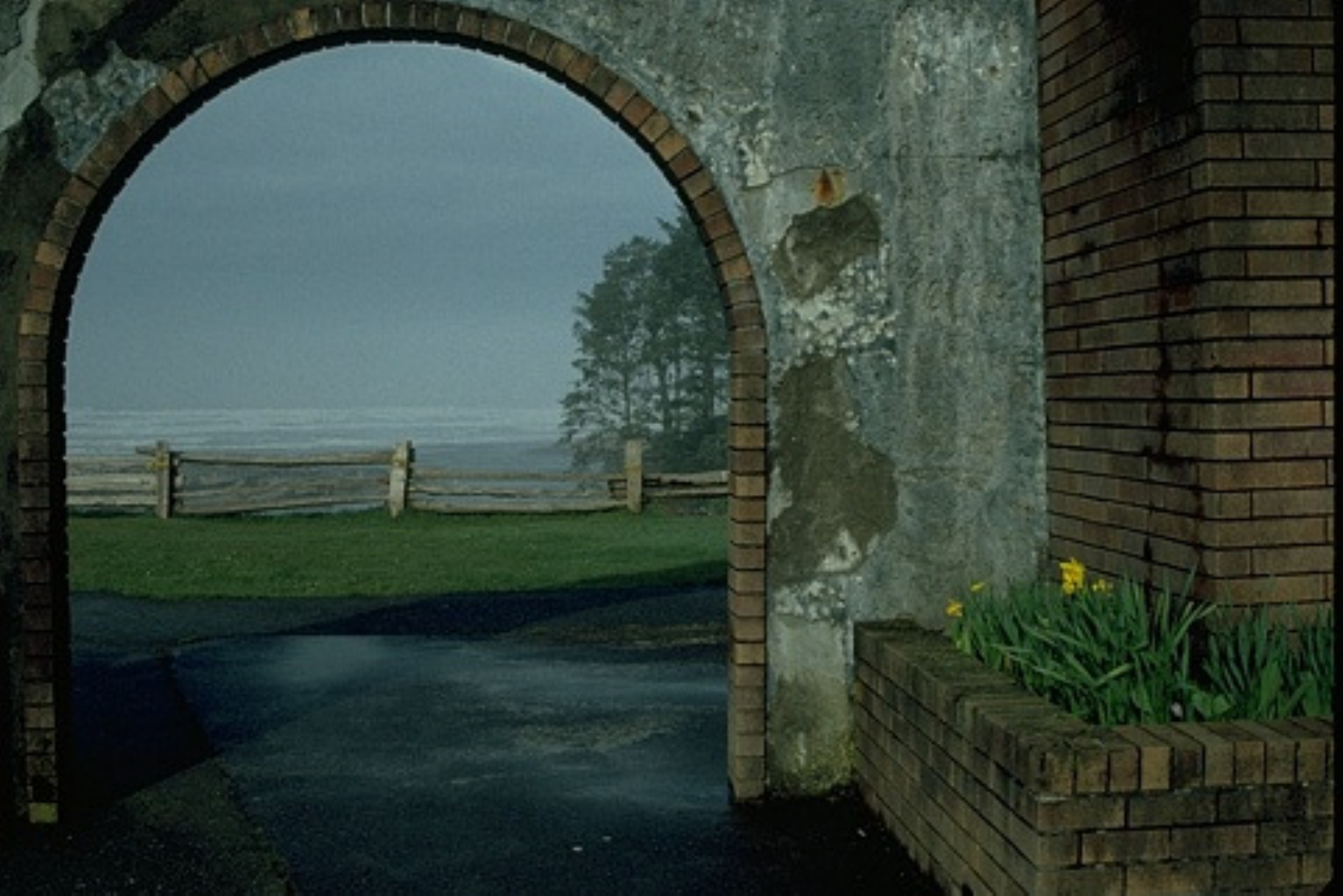}
  \end{subfigure}
  ~
  \begin{subfigure}{\threescale\linewidth}
    \centering
    \includegraphics[width=1.0\linewidth]{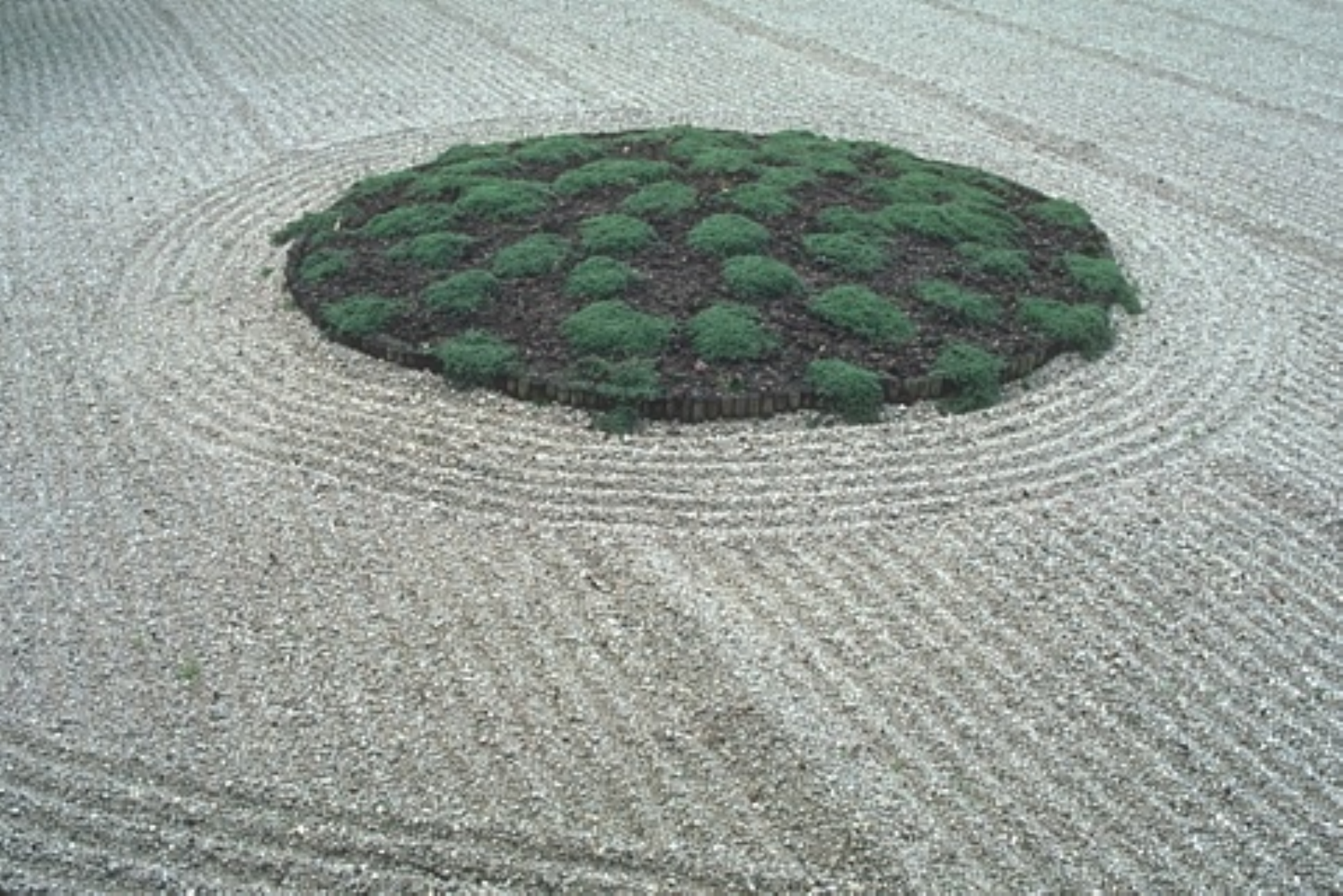}
  \end{subfigure}

  \vspace{2mm}

  \begin{subfigure}{\threescale\linewidth}
    \centering
    \includegraphics[width=1.0\linewidth]{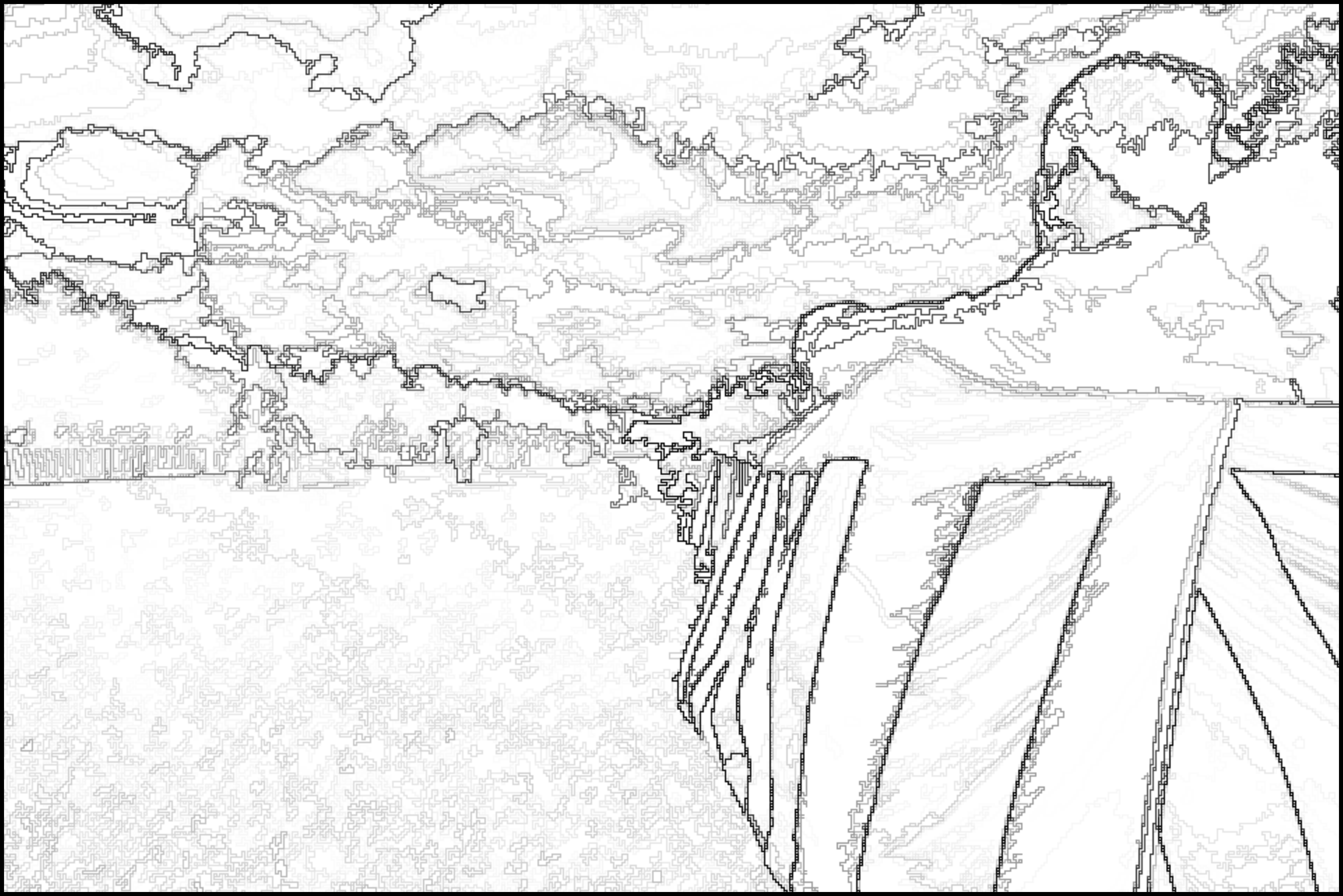}
  \end{subfigure}
  ~
  \begin{subfigure}{\threescale\linewidth}
    \centering
    \includegraphics[width=1.0\linewidth]{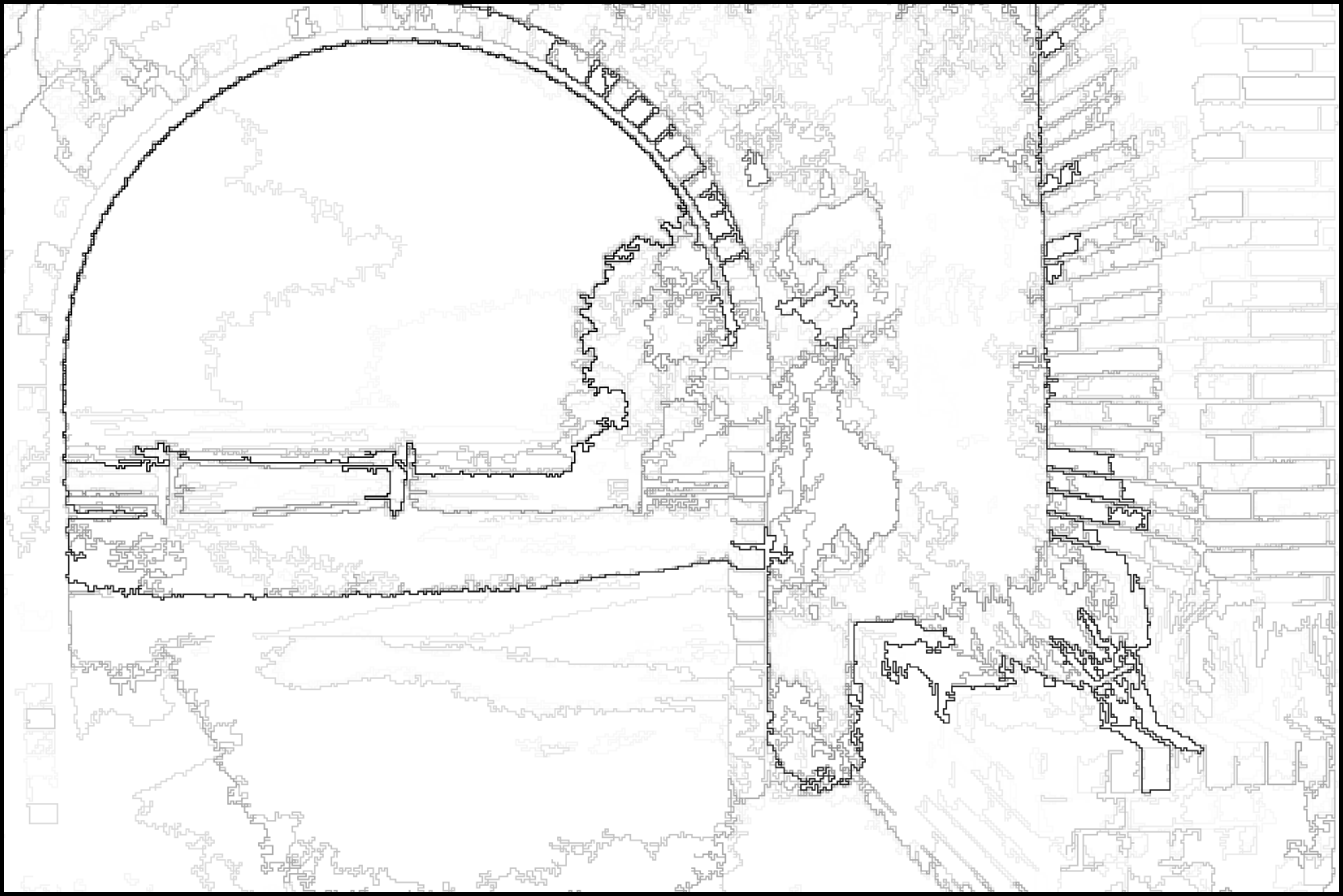}
  \end{subfigure}
  ~
  \begin{subfigure}{\threescale\linewidth}
    \centering
    \includegraphics[width=1.0\linewidth]{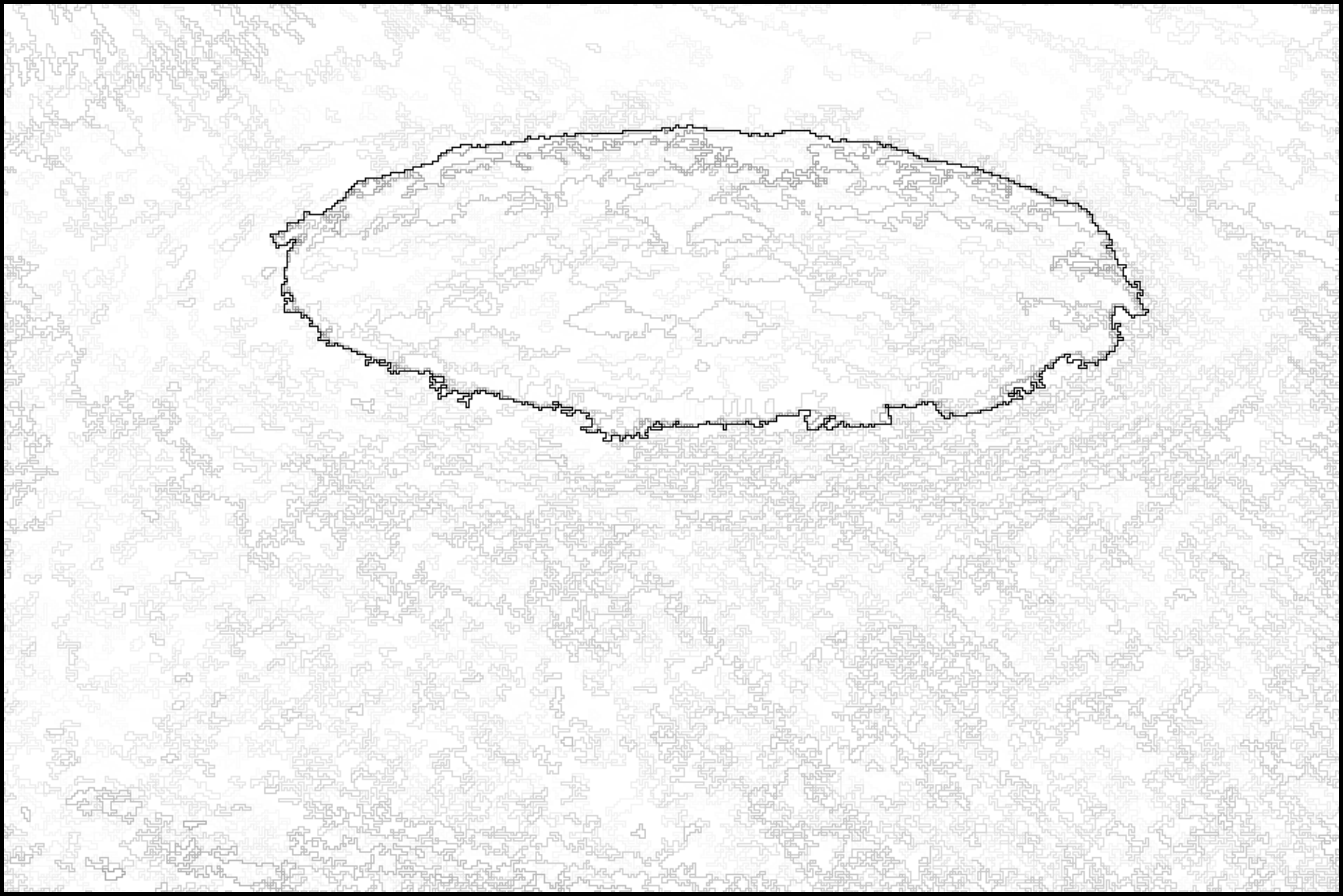}
  \end{subfigure}

  \vspace{2mm}

  \begin{subfigure}{\threescale\linewidth}
    \centering
    \includegraphics[width=1.0\linewidth]{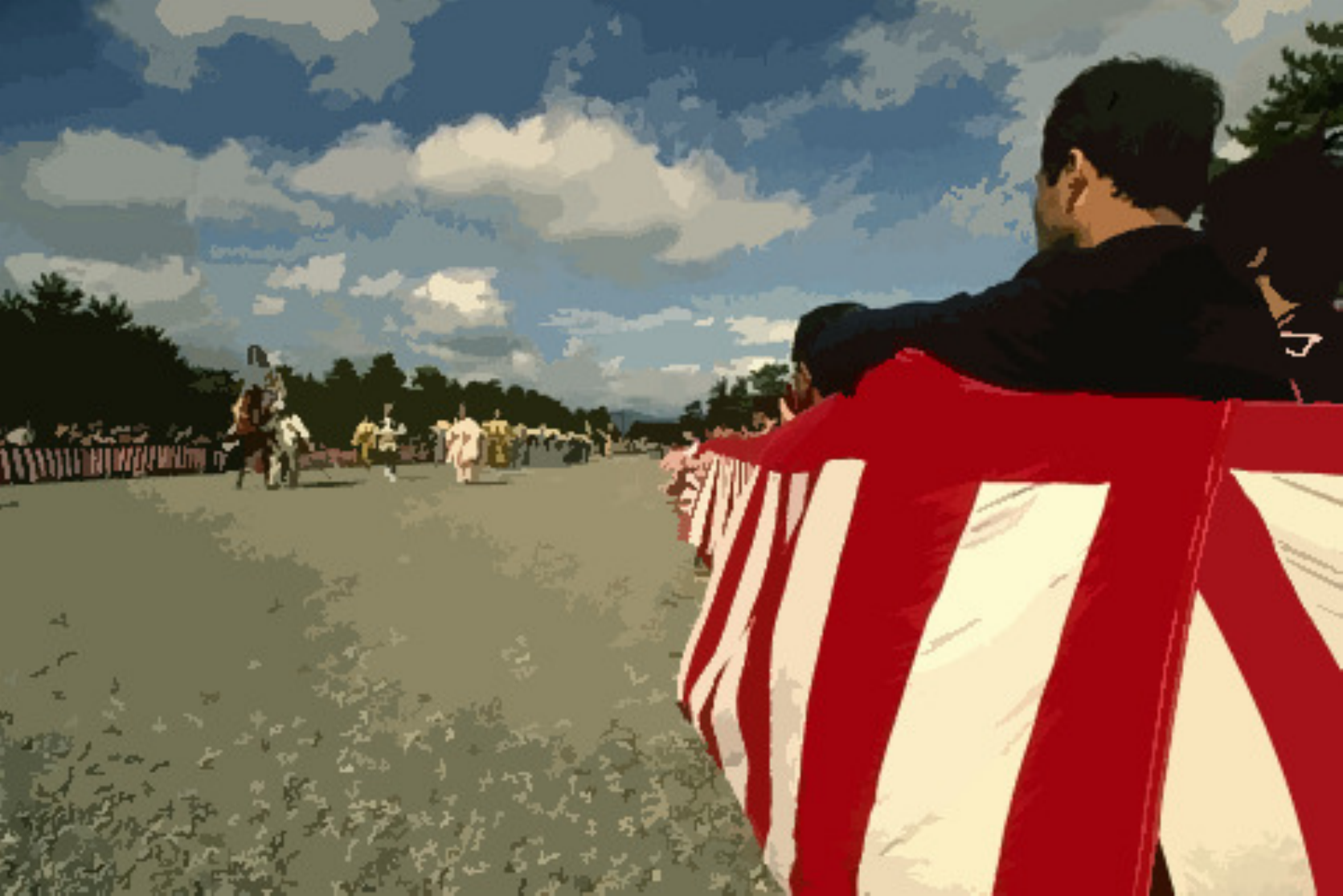}
  \end{subfigure}
  ~
  \begin{subfigure}{\threescale\linewidth}
    \centering
    \includegraphics[width=1.0\linewidth]{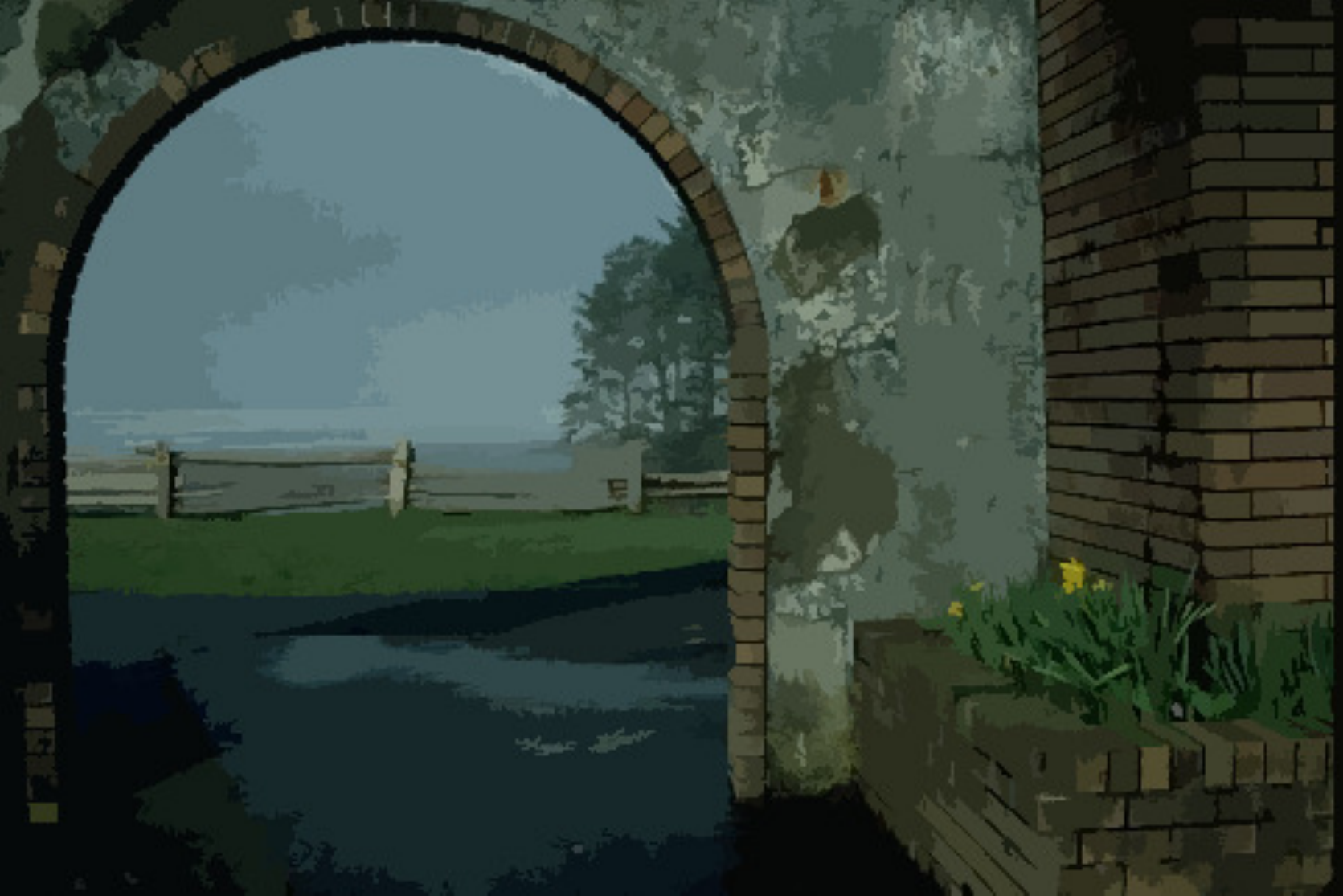}
  \end{subfigure}
  ~
  \begin{subfigure}{\threescale\linewidth}
    \centering
    \includegraphics[width=1.0\linewidth]{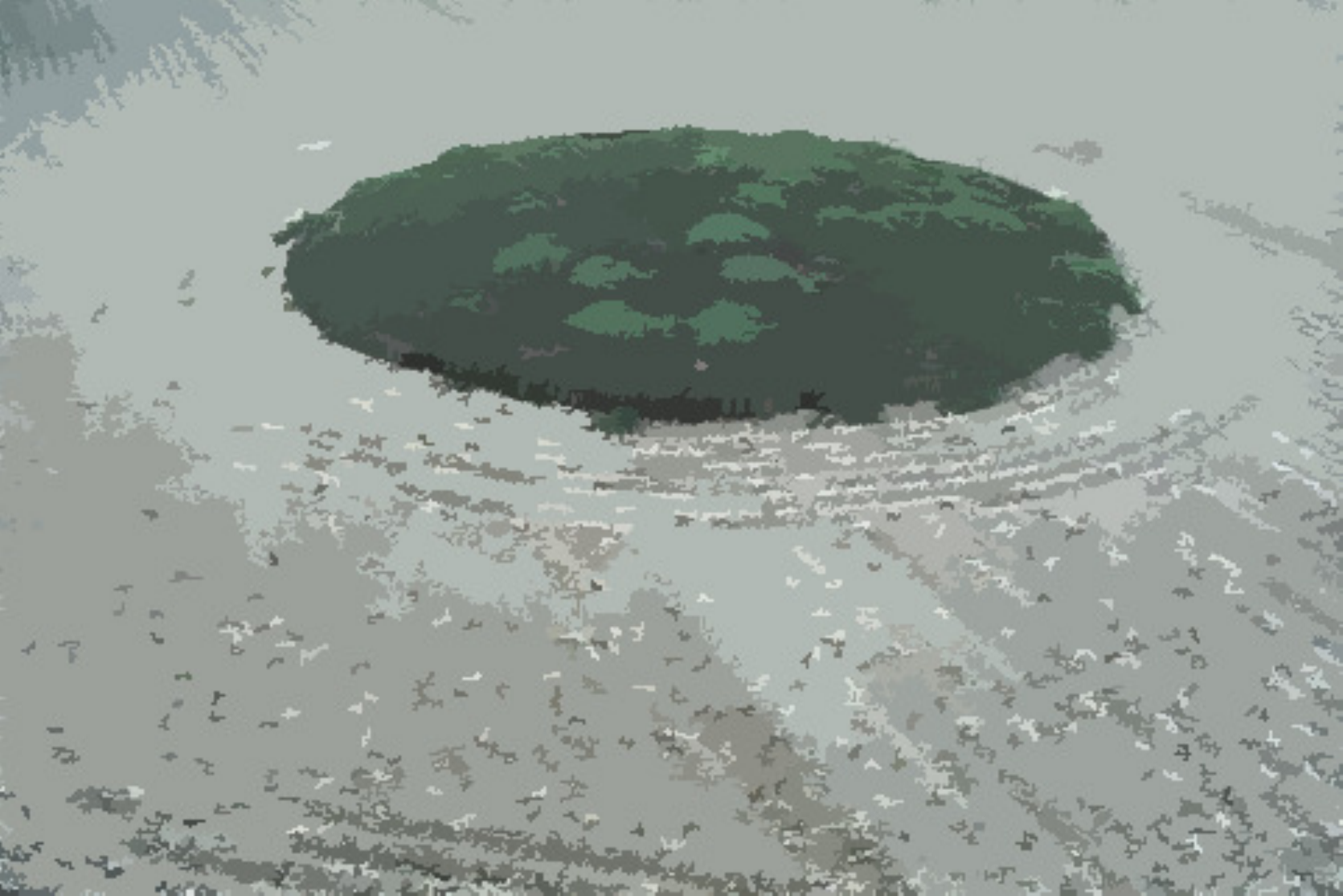}
  \end{subfigure}

  \vspace{2mm}

  \begin{subfigure}{\threescale\linewidth}
    \centering
    \includegraphics[width=1.0\linewidth]{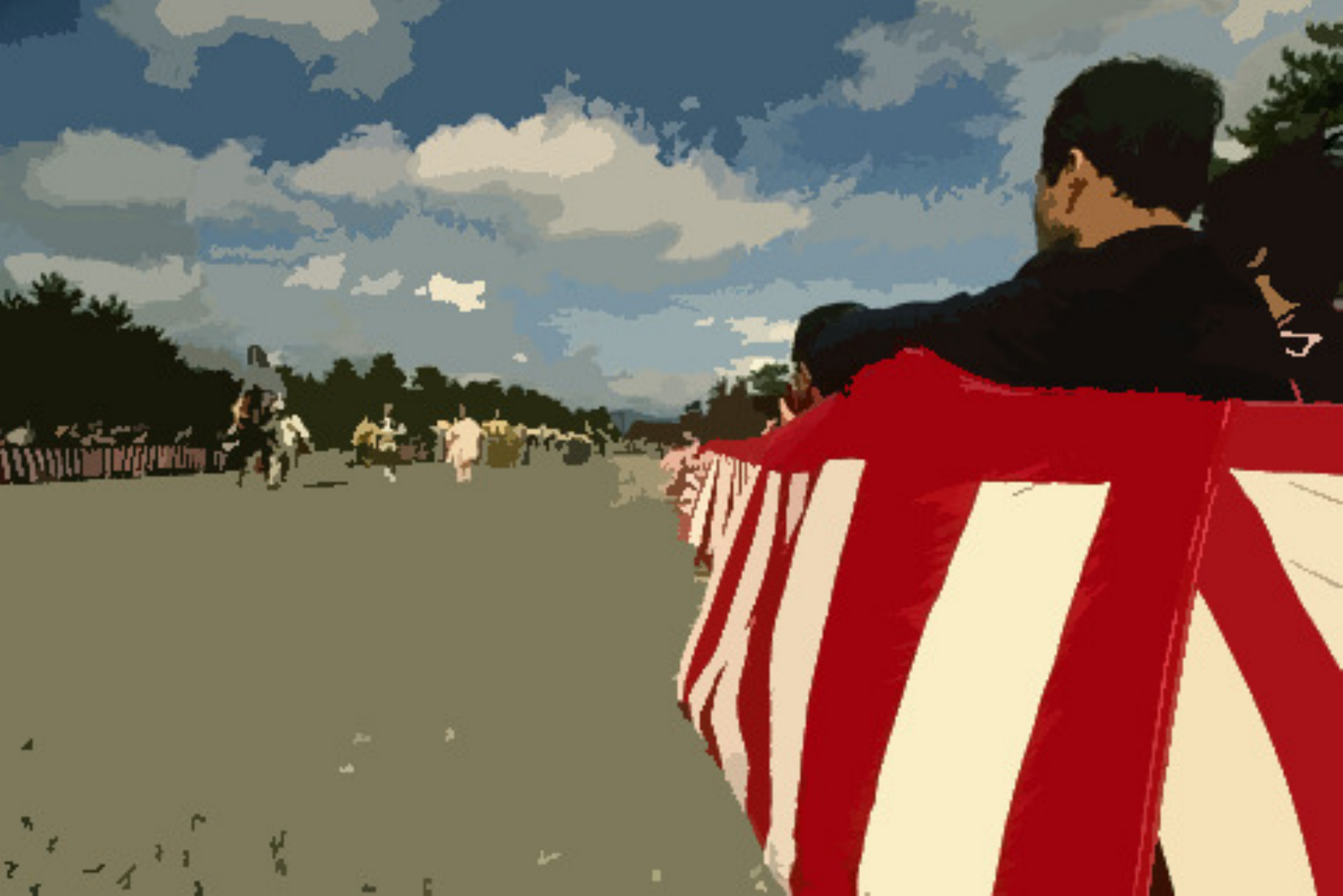}
  \end{subfigure}
  ~
  \begin{subfigure}{\threescale\linewidth}
    \centering
    \includegraphics[width=1.0\linewidth]{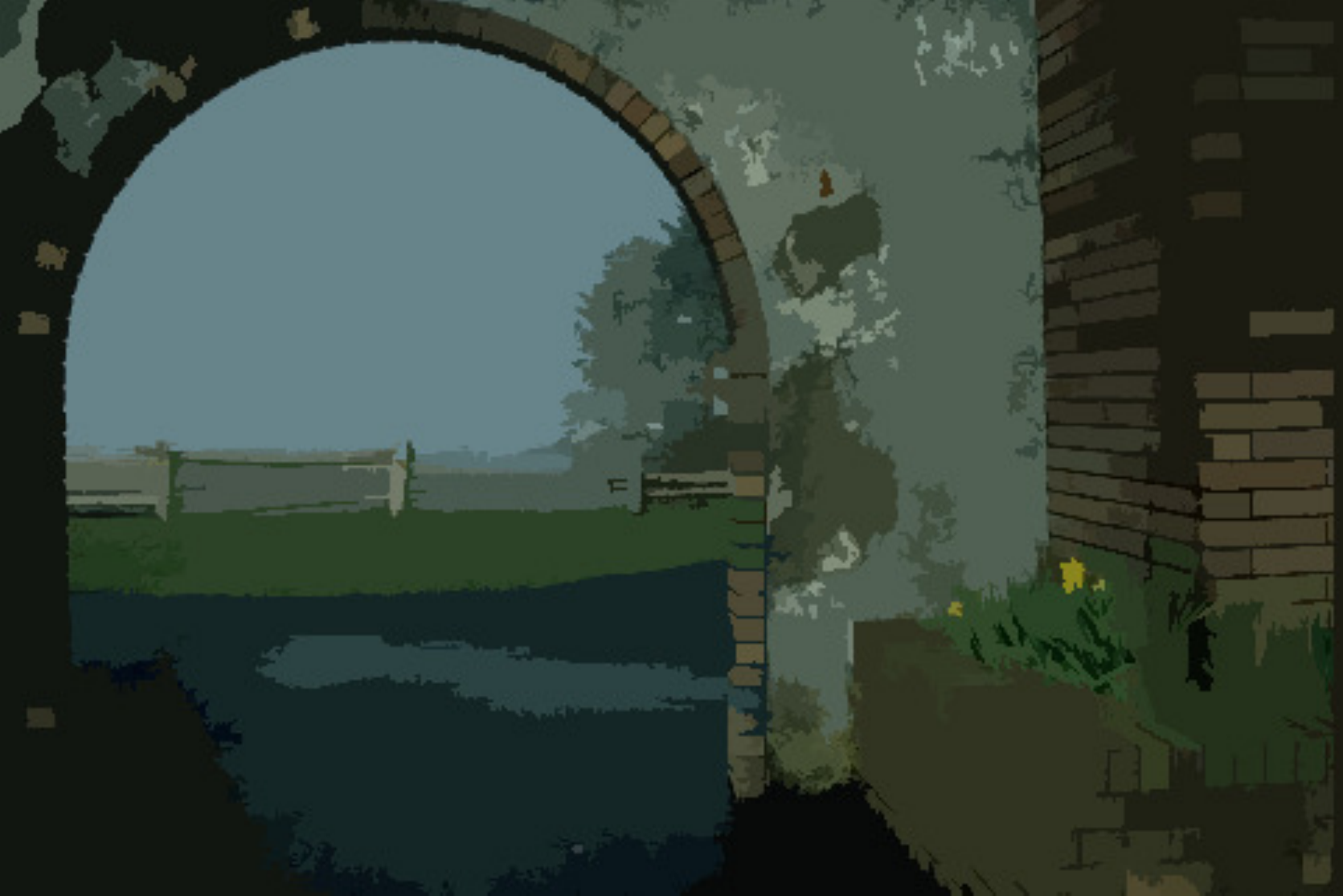}
  \end{subfigure}
  ~
  \begin{subfigure}{\threescale\linewidth}
    \centering
    \includegraphics[width=1.0\linewidth]{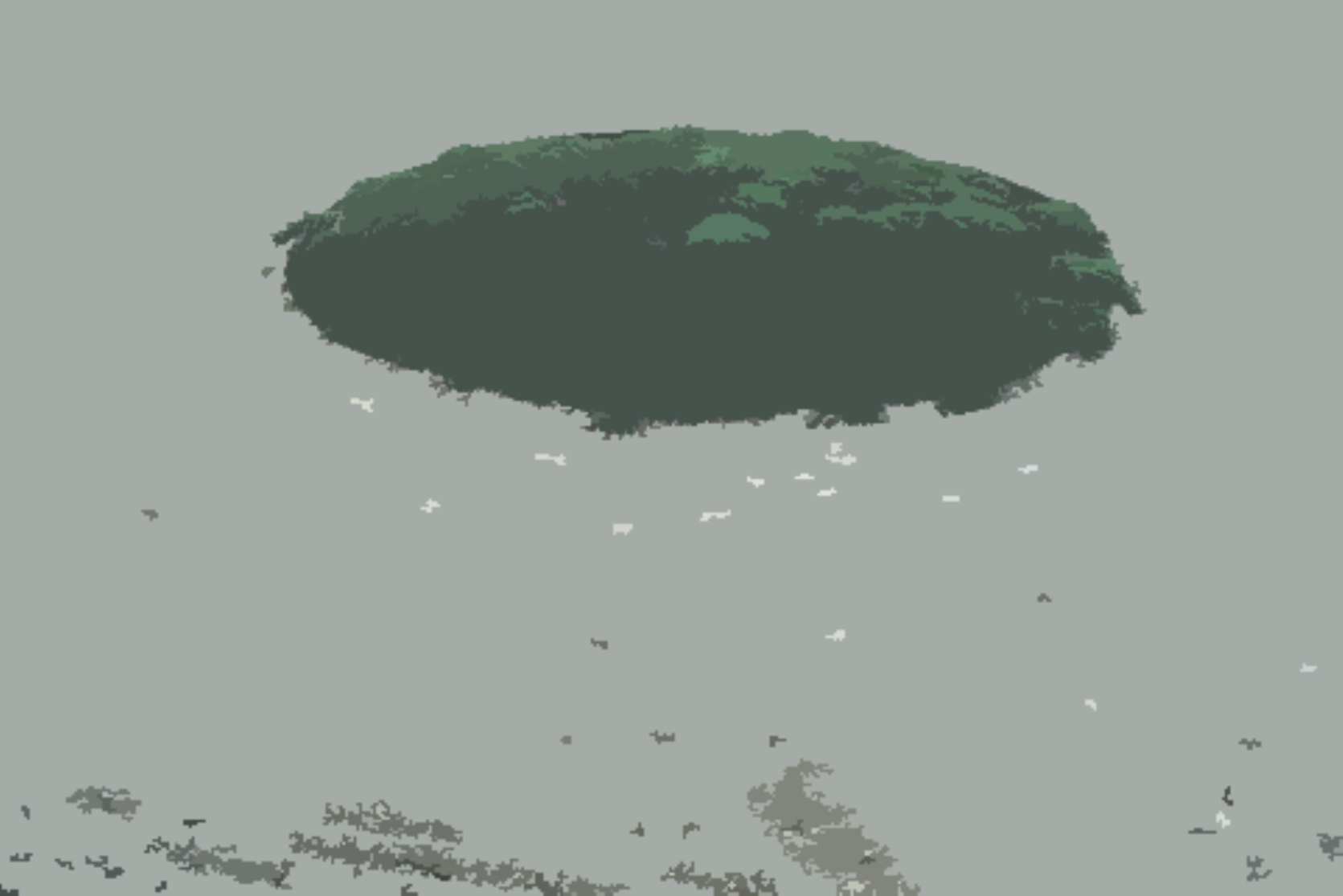}
  \end{subfigure}

  \vspace{2mm}

  \begin{subfigure}{\threescale\linewidth}
    \centering
    \includegraphics[width=1.0\linewidth]{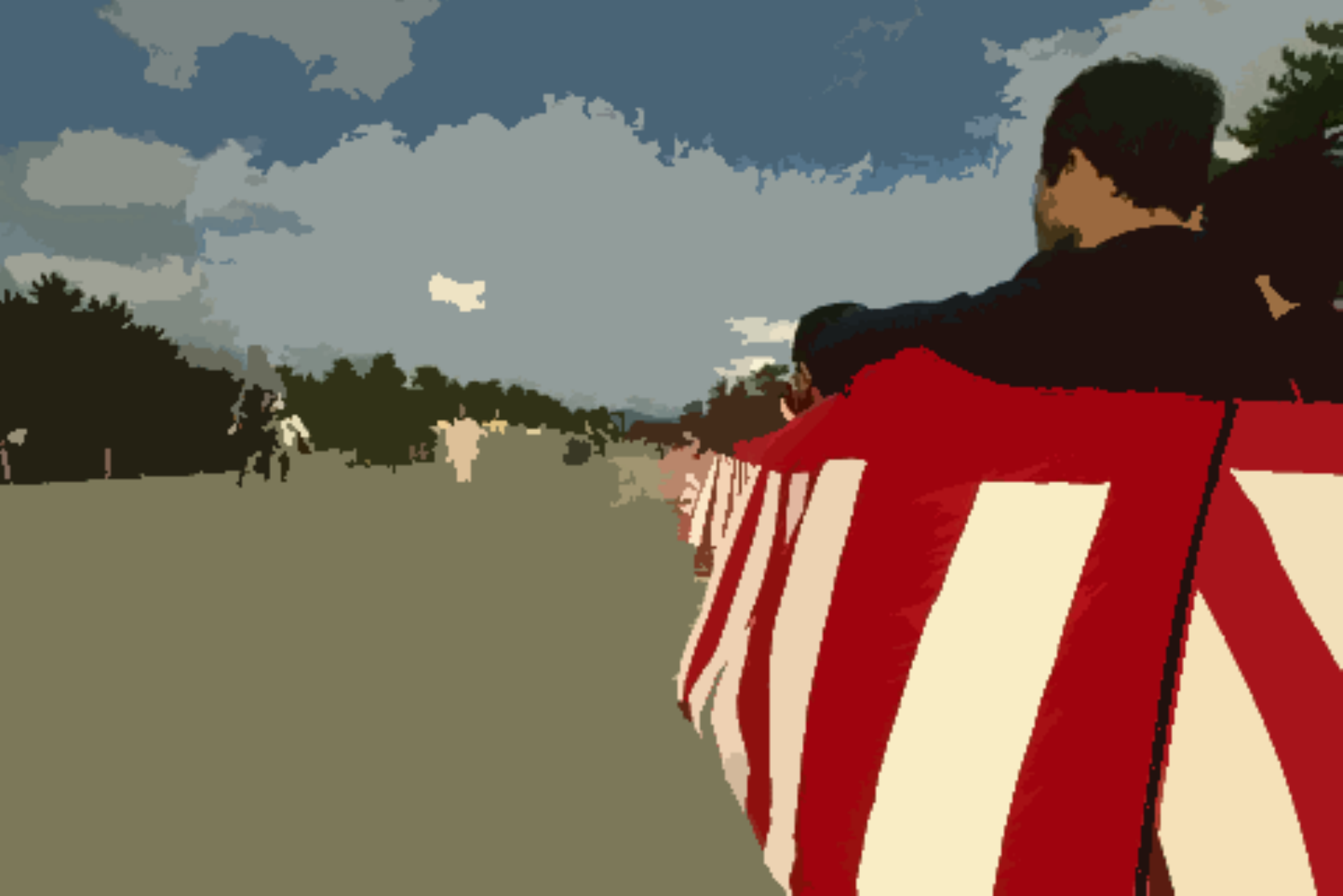}
  \end{subfigure}
  ~
  \begin{subfigure}{\threescale\linewidth}
    \centering
    \includegraphics[width=1.0\linewidth]{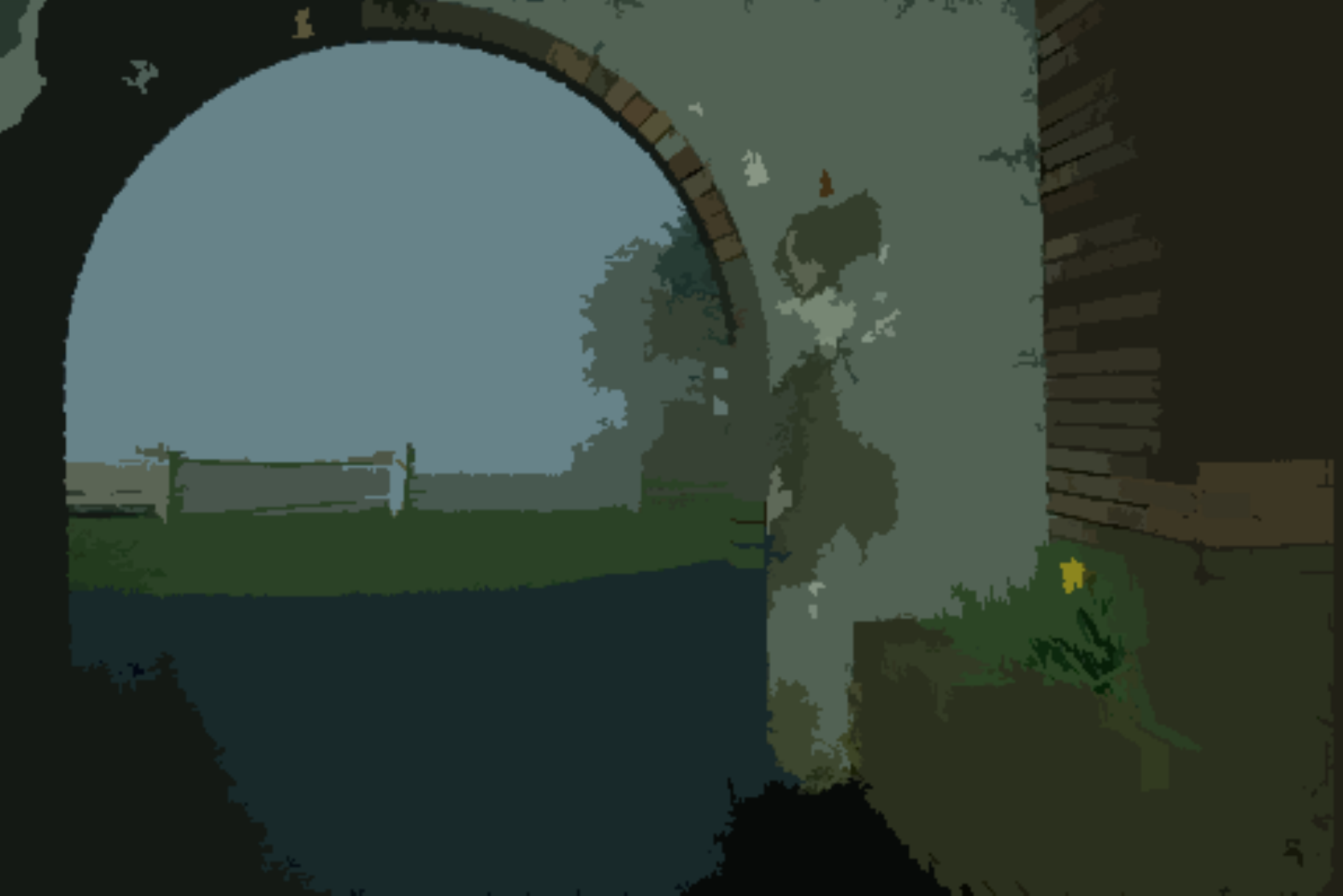}
  \end{subfigure}
  ~
  \begin{subfigure}{\threescale\linewidth}
    \centering
    \includegraphics[width=1.0\linewidth]{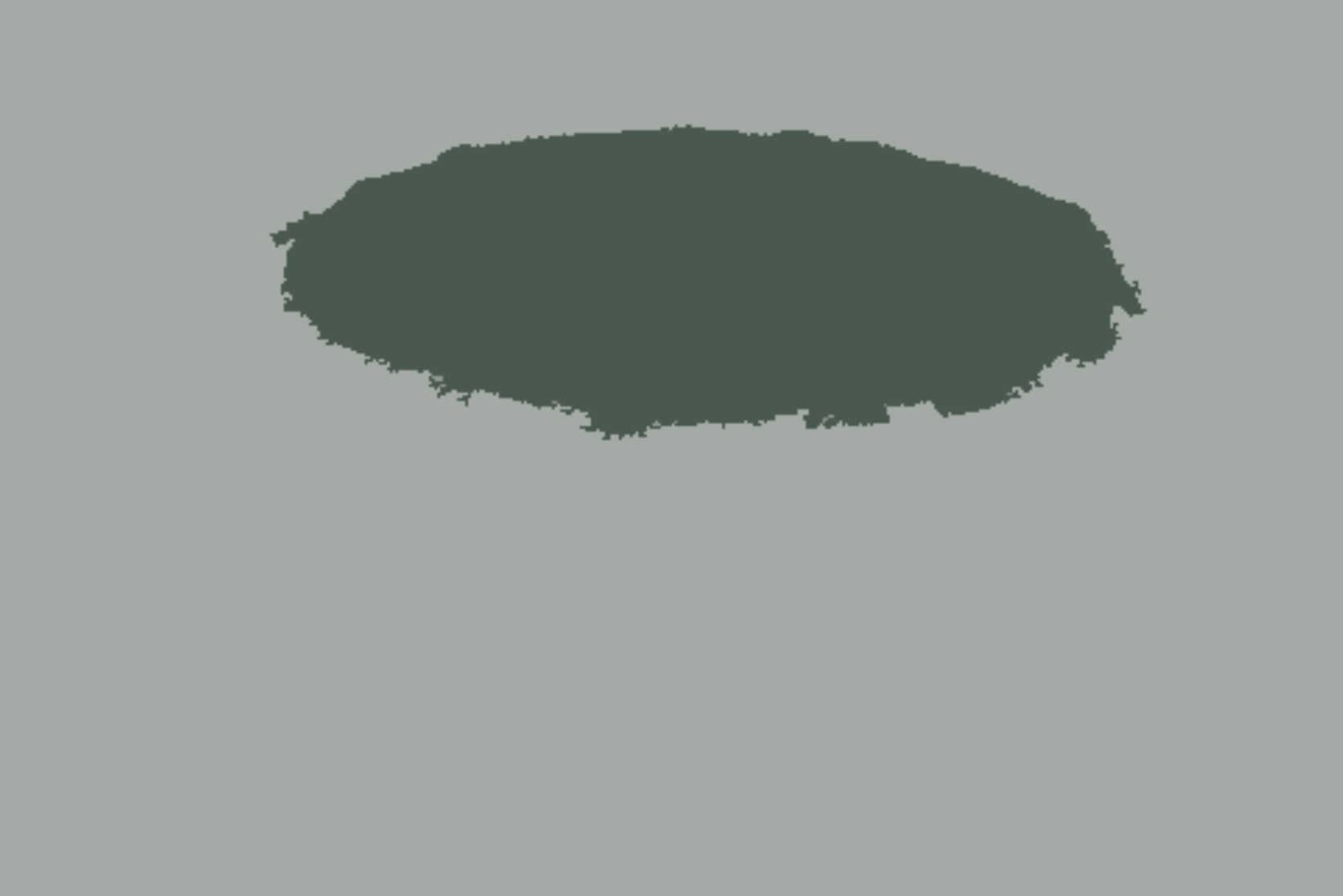}
  \end{subfigure}
  \caption{Illustration of the hierarchical image simplification using
    the attribute function $\Attribute_{\lambda_s}^\downarrow$,
    applied on some images from the dataset of BSDS500 proposed
    by~\cite{arbelaez.11.pami}. From top to bottom: input images;
    inverted saliency maps; slight simplification; moderate
    simplification; strong simplification.}
  \label{fig:hierarchicalsimplify}
\end{figure}

We have also tested our method on some cellular images, where the
method is applied on the color input image $f$. As illustrated in
Fig.~\ref{fig:cellular}, the cellular image is strongly simplified,
which almost leads to a uniform background. Finding an actual cellular
segmentation result would become much easier.

\newcommand{\fivescale}{0.18}
\begin{figure*}
  \centering
  \begin{subfigure}[b]{\fivescale\linewidth}
    \centering
    \includegraphics[width=1.0\linewidth]{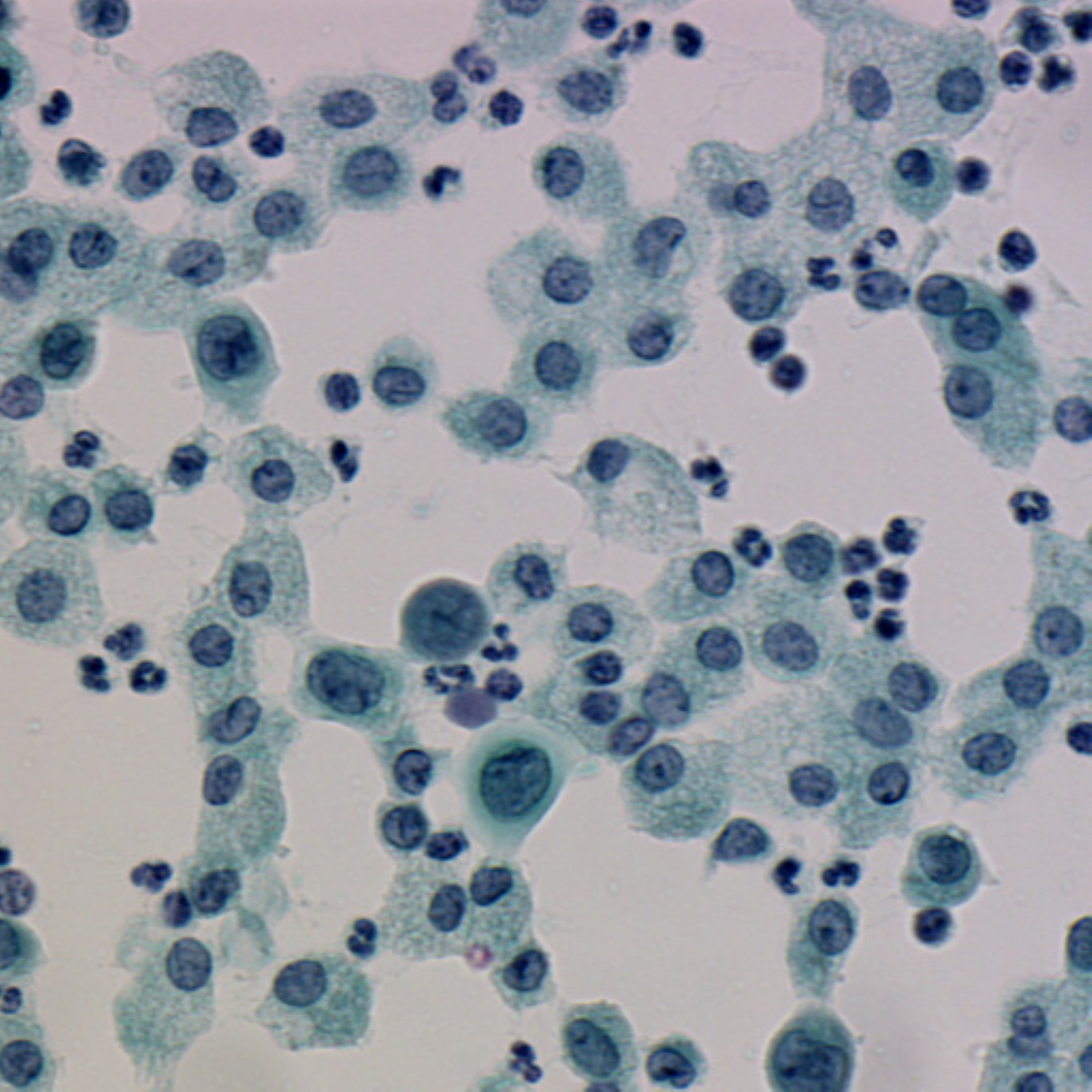}
    \caption{Input image.}
  \end{subfigure}
  ~
  \begin{subfigure}[b]{\fivescale\linewidth}
    \centering
    \includegraphics[width=1.0\linewidth]{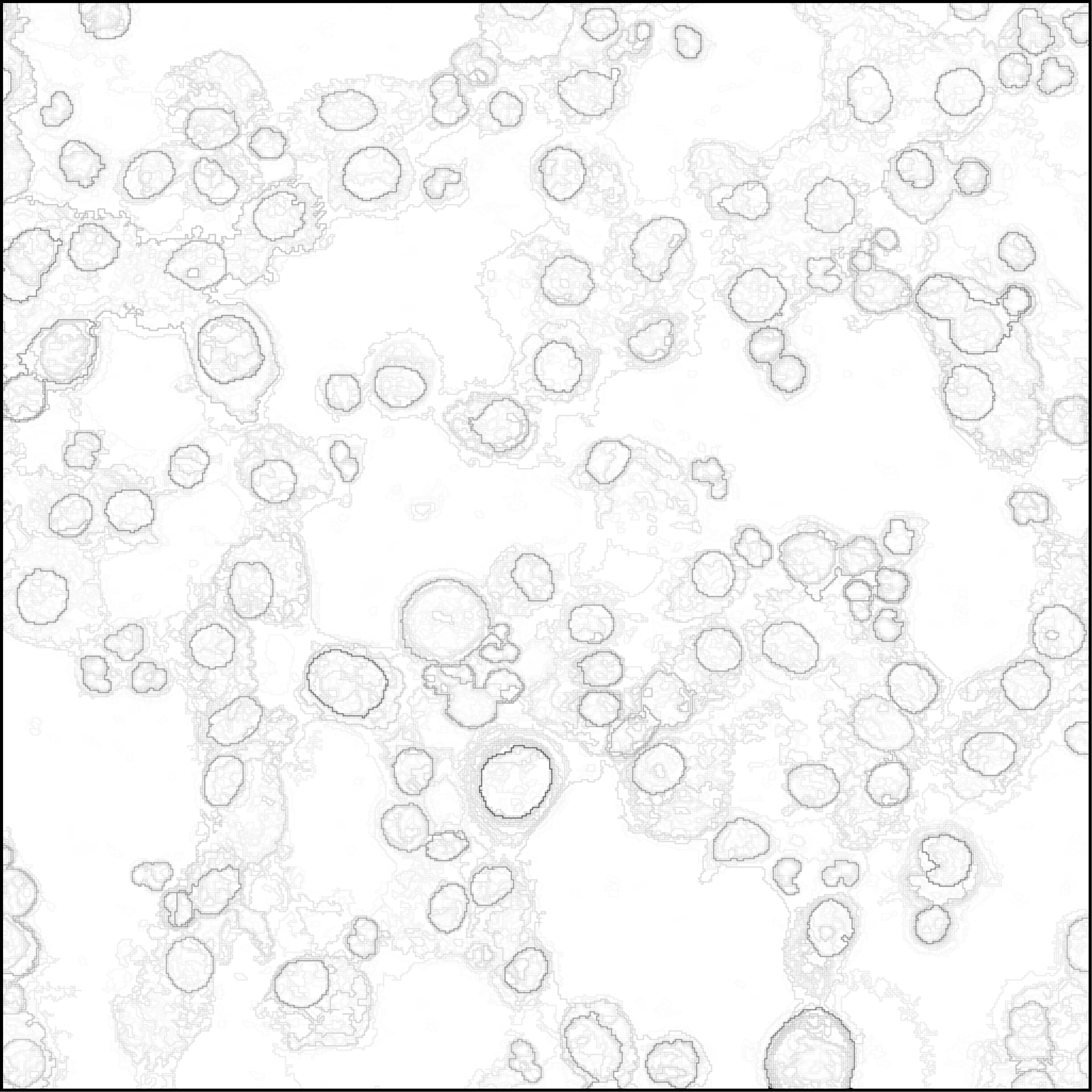}
    \caption{Saliency map.}
  \end{subfigure}
  ~
  \begin{subfigure}[b]{\fivescale\linewidth}
    \centering
    \includegraphics[width=1.0\linewidth]{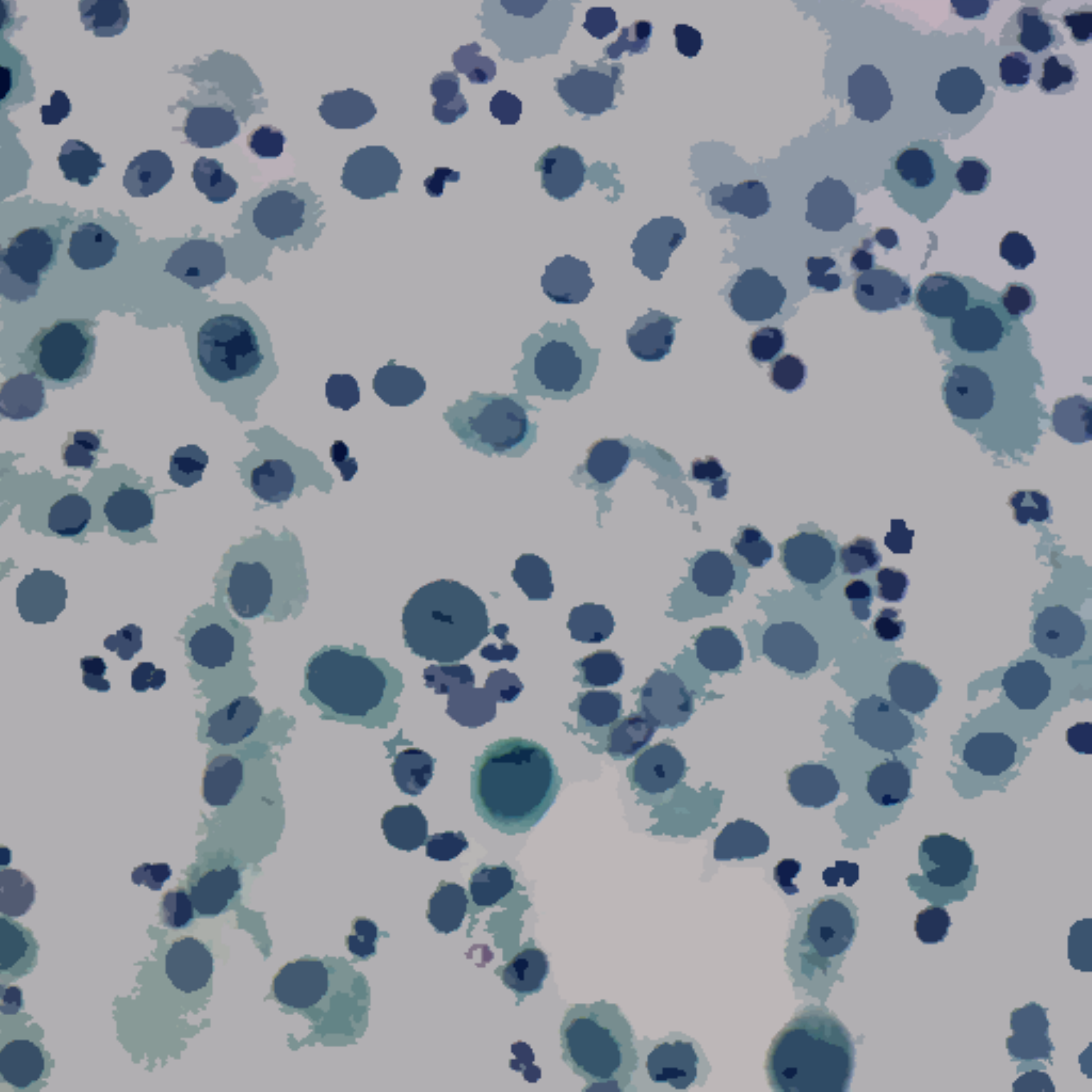}
    \caption{Slightly simplified.}
  \end{subfigure}
  ~
  \begin{subfigure}[b]{\fivescale\linewidth}
    \centering
    \includegraphics[width=1.0\linewidth]{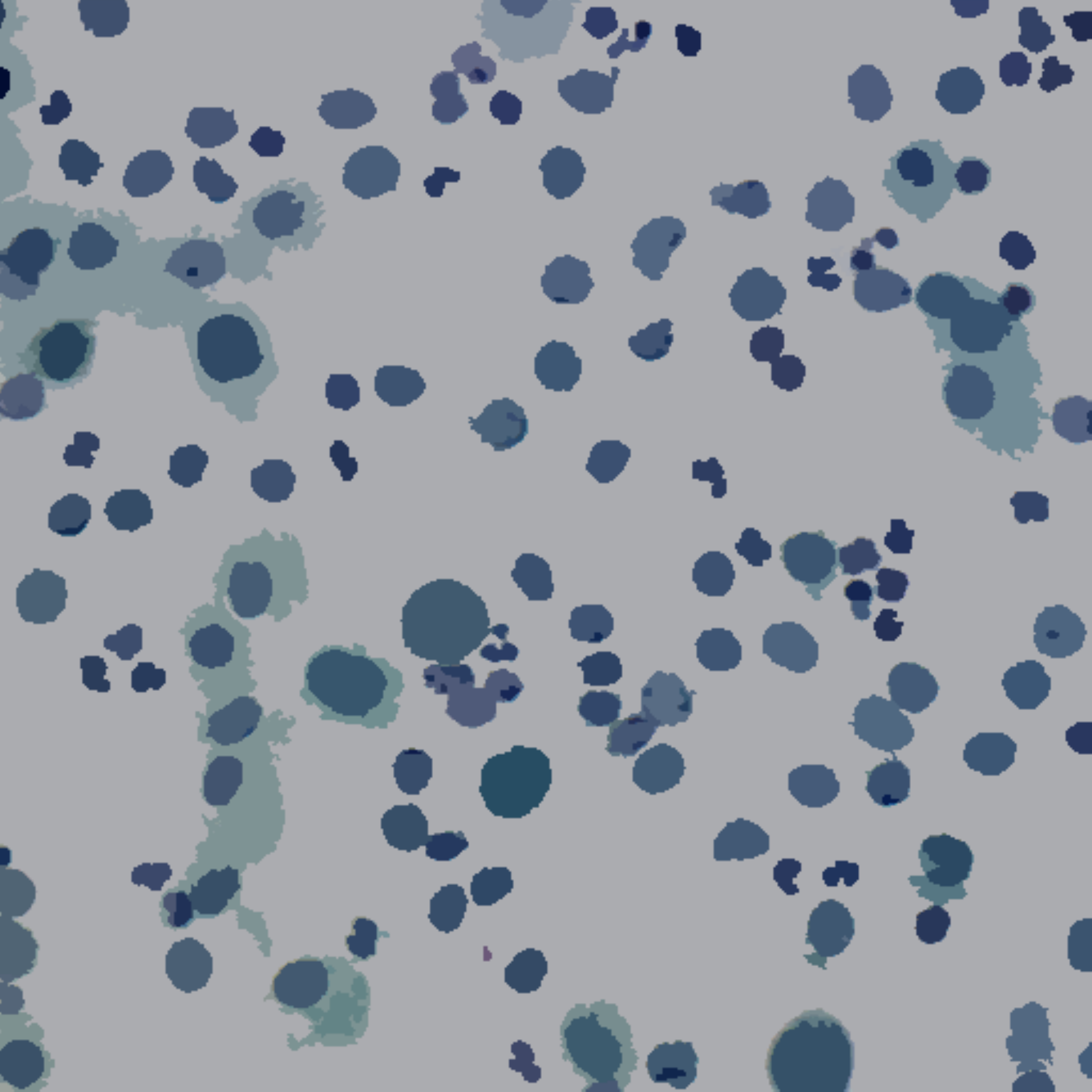}
    \caption{Moderately simplified.}
  \end{subfigure}
  ~
  \begin{subfigure}[b]{\fivescale\linewidth}
    \centering
    \includegraphics[width=1.0\linewidth]{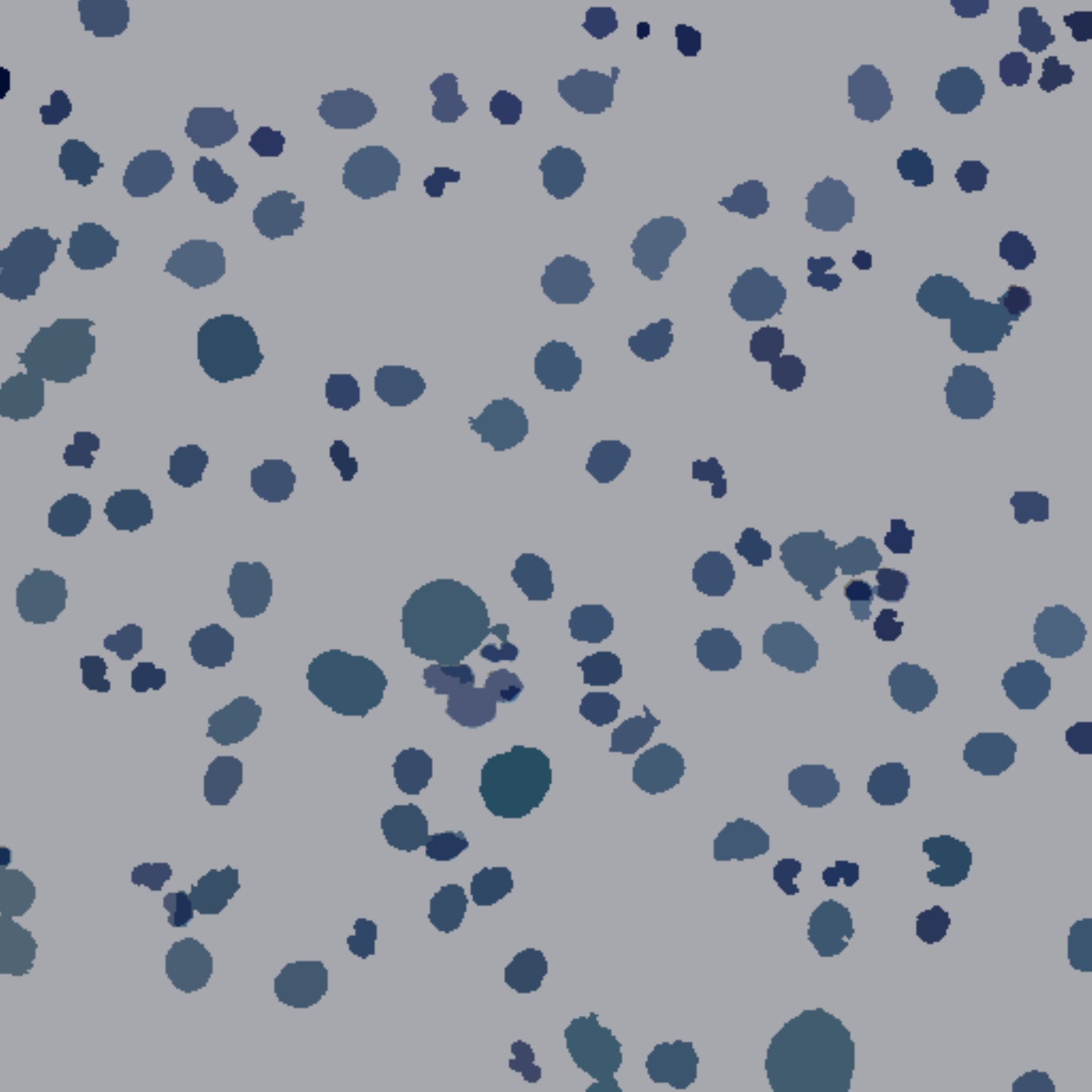}
    \caption{Strongly simplified.}
  \end{subfigure}
  \caption{Illustration of the proposed hierarchical pre-segmentations
    on a cellular image.}
  \label{fig:cellular}
\end{figure*}

\subsection{Evaluation in context of segmentation}
\label{subsec:segmentation}

We have also evaluated our hierarchical image simplifications in
context of segmentation on Weizmann segmentation evaluation database
in~\cite{alpert.12.pami}. For the 100 images containing 2 objects in
this database (See Fig.~\ref{fig:simpcolor} for several examples), the
saliency maps are thresholded with a fixed thresholding value to yield
a partition result. And we filter out the regions whose area is less
than 100 pixels. Note that, in order to perform a fair comparison with
the state-of-the art, the saliency maps are constructed using
grayscale tree of shapes computed by~\cite{geraud.13.ismm} on
grayscale versions of the input images $f$. We performed two tests as
presented in~\cite{alpert.12.pami} based on F-measure and number of
fragments. For a segmentation \textit{Seg} and a ground truth
segmentation of the object \textit{GT}, the F-measure is defined by $F
= 2 \times precision \times recall/(precision + recall)$, where
$precision = |\textit{Seg} \cap \textit{GT}|/|\textit{Seg}|$, $recall
= |\textit{Seg} \cap \textit{GT}|/|\textit{GT}|$. The number of
fragments is the number of regions selected from a partition to form
the object segmentation result \textit{Seg}.

In the first test, for each foreground object, we select the segment
that fits it the best based on F-measure score. The results of this
single segment coverage test is depicted in
Table~\ref{tab:singlesegment} (See ~\cite{alpert.12.pami} for
implementation details on the settings for the other methods).  In
this test, our method achieves F-measure score on par with the
state-of-the-art methods, especially when replacing, in the attribute
$\Attribute_\nabla$, the classical grayscale gradient with the
(learned) gradient computed (on the grayscale input image $f$)
by~\cite{dollar.15.pami} (see ``Our2''). By using another gradient, we
change the order in which the nodes of the tree are processed; thus
this result highlights the importance of the sorting step in our
algorithm. In Table~\ref{tab:singlesegment}, note that {\em
  Gpb-owt-ucm without texture} denotes the method of Gpb-owt-ucm
computed without taking into account the texture information in the
Gpb part.  More precisely, in this case, the Gpb is computed using
only brightness and color gradients. Note also that our method does
not explicitely use any texture information either.

In the second test, a combination of segments whose area overlaps
considerably the foreground objects is utilized to assess the
performance. For each union of segments, we measure the F-measure
score and the number of segments composing it. This test is a
compromise between good F-measure and low number of fragments.  The
results of this fragmentation test is given in
Table~\ref{tab:multisegment}. In this test the averaged F-measure of
different methods is fairly similar. However, our method, as a
pre-segmentation method without using any texture information has a
relatively high number of fragments.

The saliency maps for these images containing two objects in the
Weizmann dataset is available on
\url{http://publications.lrde.epita.fr/xu.hierarchymsll}.

\begin{table}[ht]
  \caption{Results of single segment coverage test using F-measure on
    the two objects dataset in~\cite{alpert.12.pami}.  ``Our2'' stands
    for our method with the attribute $\Attribute_\nabla$ using the
    gradient computed by~\cite{dollar.15.pami}.}
  \label{tab:singlesegment}
  \centering
  \begin{tabular}{|c|c|c|c|}
    \hline
    \small{Method} & \small{Average} & \small{Larger} & \small{Smaller}\\
    \hline
    \small{\it Our2} & \small{\it \textbf{0.80}} & \small{\it \textbf{0.79}} & \small{\it \textbf{0.81}} \\
    \hline
    \small{\textrm{Gpb-owt-ucm 
    }} & \small{\textbf{0.79}} & \small{\textbf{0.80}} & \small{\textbf{0.78}} \\
    \hline
    \small{\it Our} & \small{\it \textit{0.77}} & \small{\it \textit{0.77}} & \small{\it \textit{0.76}} \\
    \hline
    \small{Gpb-owt-ucm without texture} & \small{0.76} & \small{0.76} & \small{0.75} \\
    \hline
    \small{Gpb in~\cite{alpert.12.pami}} & \small{0.72} & \small{0.70} & \small{0.75} \\
    \hline
    \small{Method in~\cite{alpert.12.pami}} & \small{0.68} & \small{0.70} & \small{0.65} \\
    \hline
    \small{SWA in~\cite{sharon.06.nature}} & \small{0.66} & \small{0.74} & \small{0.57} \\
    \hline
    \small{MeanShift 
    } & \small{0.61} & \small{0.65} & \small{0.58} \\
    \hline
    \small{N-Cuts} 
    & \small{0.58} & \small{0.66} & \small{0.49} \\
    \hline
  \end{tabular}
\end{table}
\begin{table*}
  \caption{Fragmented coverage test results for the two objects
    dataset proposed by~\cite{alpert.12.pami}: compromise between good
    F-measure and low number of fragments. Our results are comparable
    to the state-of-the-art. ``Our2'' stands for our method with the
    attribute $\Attribute_\nabla$ using the gradient computed
    by~\cite{dollar.15.pami}.}
  \label{tab:multisegment}
  \centering
  \begin{tabular}{|c|c|c|c|c|c|c|}
    \hline
    \multirow{2}{*}{\small{Method}} & \multicolumn{2}{|c|}{\small{Averaged}} & \multicolumn{2}{|c|}{\small{Larger object}} & \multicolumn{2}{|c|}{\small{Smaller object}}\\
     \cline{2-7} & \small{F-measure} & \small{\#fragments} & \small{F-measure} &  \small{\#fragments}  & \small{F-measure} & \small{\#fragments}\\
    \hline
    \small{SWA in~\cite{sharon.06.nature}} & \small{\bf 0.88} & \small{3.13} & \small{\bf 0.91} & \small{3.88} & \small{\bf 0.84} & \small{2.37} \\
    \hline
    \small{\it Our2} & \small{\it \textbf{0.86}} & \small{\it 2.40} & \small{\it 0.85} & \small{\it 3.00} & \small{\it \textbf{0.86}} & \small{\it 1.81} \\
    \hline
    \small{\textrm{Method in~\cite{alpert.12.pami}}} & \small{\textrm{0.85}} & \small{\textbf{1.67}} & \small{0.87} & \small{\textbf{2.00}} & \small{\textbf{0.84}} & \small{\textbf{1.33}} \\
    \hline
    \small{N-Cuts in~\cite{shi.00.pami}} & \small{0.84} & \small{2.64} & \small{\bf 0.88} & \small{3.34} & \small{0.80} & \small{1.93} \\
    \hline
    \small{Gpb reported in~\cite{alpert.12.pami}} & \small{0.84} & \small{2.95} & \small{0.87} & \small{3.60} & \small{0.81} & \small{2.30} \\
    \hline
    \small{\it Our} & \small{\it 0.83} & \small{\it 3.16} & \small{\it 0.85} & \small{\it 4.10} & \small{\it 0.81} & \small{\it 2.23} \\
    \hline
    \small{Gpb-owt-ucm in~\cite{arbelaez.11.pami}} & \small{0.82} & \small{\bf 1.57} & \small{0.84} & \small{\bf 1.79} & \small{0.81} & \small{\bf 1.35} \\
    \hline
    \small{Gpb-owt-ucm without texture} &  \small{0.81} & \small{2.72} & \small{0.82} & \small{3.32} & \small{0.80} & \small{2.12}\\
    \hline
    \small{MeanShift in~\cite{comaniciu.02.pami}} & \small{0.78} & \small{3.65} & \small{0.85} & \small{4.49} & \small{0.71} & \small{2.81} \\
    \hline
  \end{tabular}
\end{table*}
%



\section{Comparison with similar works}
\label{sec:relatedwork}
The tree of shapes has been widely used in connected operators,
filtering tools that act by merging flat zones for image
simplification and segmentation.
The simplification and segmentation relies on relevant shapes
extraction (\ie, salient level lines), usually achieved by tree
filtering based on some attribute function. A detailed review of tree
filtering strategies can be found in~\cite{salembier.09.spm}. In all
these strategies, the attribute function $\Attribute$ characterizing
each node plays a very important role in connected filtering.  The
classical connected operators make filtering decisions based only on
attribute function itself or the inclusion relationship of the tree
(\eg,~\cite{xu.14.filter}). They are usually performed by removing the
nodes whose attributes are lower than a given threshold. The method we
propose in this paper combines this idea of classical connected
operators with the energy minimization problem of
Eq~(\ref{eq:solution}). It also makes use of the spatial information
of the original image from which the tree is constructed. This might
give more robust filtering decision.

In this paper, we focus particularly on hierarchical relevant shapes
selection by minimizing some multiscale affine separable energy
functional (\eg, piecewise-constant Mumford-Shah functional). The
closely related work is the one in~\cite{guigues.06.ijcv}, where the
authors proposed the scale-set theory, including an efficient greedy
algorithm to minimize this kind of energy on a hierarchy of
segmentations. More precisely, the authors use dynamic programming to
efficiently compute two scale parameters $\lambda_s^+$ and
$\lambda_s^-$ for each region $R$ of the input hierarchy $H$, where
$\lambda_s^+$ (\resp, $\lambda_s^-$) corresponds to the smallest
parameter $\lambda_s$ such that the region $R \in H$ belongs
(\resp. does not belong) to the optimal solution of segmentation by
minimizing $E_{\lambda_s}$, we have $\lambda_s^-(R) = \underset{R' \in
  H, R \subset R'}{\mathrm{min}}\lambda_s^+(R')$.  There may exist
some regions $R$ such that $\lambda_s^-(R) \leq \lambda_s^+(R)$, which
implies that the region $R \in H$ does not belong to any optimal cut
of $H$ by minimizing the energy $E_{\lambda_s}$. One removes these
regions from the hierarchy $H$ and updates the parenthood relationship
which yields a hierarchy $H'$, a hierarchy of global optimal
segmentations on the input hierarchy. This work has been continued and
extended by~\cite{kiran.14.pr}. These methods work on an input
hierarchy of segmentations, which is very different from the tree of
shapes (a natural and equivalent image representation). {\bluetext
  Indeed, each cut of the tree of shapes is a subset of the image
  domain, while each cut of the hierarchy of segmentations forms a
  partition of the image domain; see~\cite{ronse.14.jmiv}. This basic
  difference prohibits the direct use on the tree of shapes of the
  classical works which find optimal hierarchical segmentations by
  energy minimization.} In this sense, our approach can be seen as an
extension of the scale-set theory proposed in~\cite{guigues.06.ijcv}
to the tree of shapes.


Another related work is the one in~\cite{ballester.07.jmiv}. It also
selects meaningful level lines for image simplification and
segmentation by minimizing the piecewise-constant Mumford-Shah
functional. For this method, at each step the level line is selected
which inflicts the largest decrease of functional.  As a consequence,
the iterative process of~\cite{ballester.07.jmiv} requires not only
computing a lot of information to be able to update the functional
after each level line suppression, but also to find at each step,
among all remaining level lines, the one candidate to the next
removal. Consequently, the optimization process has a $O(N_\Tree^2)$
time complexity w.r.t. the number of nodes $N_\Tree$ of the tree. A
heap-based implementation may improve the time complexity, but since
at each removal, one has to update the corresponding energy variation
for its children, parent, siblings, maintaining the heap structure is
a costly process. In practice, the gain using heap-based
implementation is relatively insignificant.
Hence~\cite{ballester.07.jmiv} is computationally expensive. We
propose to fix that issue thanks to a reasonable ordering of level
lines based on their quantitative meaningfulness measurement
{\bluetext (\eg, the average of gradient's magnitude along the level line
  $\Attribute_\nabla$)}. The time complexity of our optimization
process is linear w.r.t. the number of nodes $N_\Tree$. We have
implemented the method of~\cite{ballester.07.jmiv} using the same tree
construction algorithm and the same data structure based on heap. We
have compared the running time on 7 classic images on a regular PC
station. The comparison is detailed in
Table.~\ref{tab:comparison}. Our proposal is significantly faster than
that of~\cite{ballester.07.jmiv}. Our approach is almost linear
w.r.t. the number of nodes in the tree. Yet, the method
of~\cite{ballester.07.jmiv} seems to depend also on the depth of the
tree. In~\cite{ballester.07.jmiv}, the authors proposed applying the
simplification scheme successively with a set of augmenting parameters
$\lambda_s$ so that to construct the input hierarchy. Then they
employed the scheme of scale-set theory proposed
by~\cite{guigues.06.ijcv} on the obtained hierarchy to achieve a final
hierarchy of optimal segmentations. In our case, rather than using a
fixed parameter $\lambda_s$ or a set of fixed parameters, we propose
to assign a measure related to $\lambda_s$ to each shape as an
attribute function. Then we use the hierarchy transformation (reviewed
in Section~\ref{subsec:ssf}) based on extinction values and on a
tree-based shape space to compute a hierarchical salient level lines
selection.

\begin{table}
  \centering
  \caption{Comparison of computation times on seven classical
    images. The size for image ``House'' and ``Camera'' is
    $256\times256$, and $512\times512$ for the other images.}
  \label{tab:comparison}
  \begin{tabular}{|c|c|c|c|c|c|}
    \hline
    \multirow{2}{*}{\small{Image}} & \multirow{2}{*}{\small{Depth}} & \multirow{2}{*}{\small{\#Nodes}} & \multicolumn{2}{|c|}{\small{Time (s)}}\\
     \cline{4-5}  & & &  \small{\cite{ballester.07.jmiv}} &  \small{Our}\\
    \hline
    \small{\it House} & \small{126} & \small{23588} & \small{4.11} & \small{0.22}\\
    \hline
    \small{\it Camera} & \small{126}  & \small{24150} & \small{4.19} & \small{0.23}\\
    \hline
    \small{\it Lena} & \small{141} & \small{84699} & \small{27.77} & \small{0.92}\\
    \hline
    \small{\it Peppers} & \small{176} & \small{97934} & \small{48.18} & \small{0.93}\\
    \hline
    \small{\it Boat} & \small{255} & \small{100518} & \small{87.24} & \small{0.94} \\
    \hline
    \small{\it Barbara}  & \small{131} & \small{106285} & \small{51.87} & \small{0.99}\\
    \hline
    \small{\it Mandrill} & \small{185} & \small{153029} & \small{200.22} & \small{1.34}\\ 
    \hline
  \end{tabular}
\end{table}


It is worth noticing that the minimization of Mumford-Shah-like
functional has also been applied to shape analysis
in~\cite{tari.14.jmiv}. It consists of adding a non-local term, which
is the squared average of the field in the energy functional. Its
minimization tends to form negative field values on narrow or small
parts as well as on protrusions, and positive field values on central
part(s) of the input shape. The negative and positive regions inside
the input shape yield some saddle points at which a crossing of a
level curve occurs. This leads to a binary partition hierarchy $H_b$
of the shape. Then a probability measure based on the obtained field
values inside the shape is assigned to each node of the partition
hierarchy $H_b$. A set of hierarchical representations of the shape is
obtained by removing some nodes from $H_b$ and update the parenthood
relationship. Each candidate hierarchical representation is assigned
with a saliency measure given by the products of the probability
measure of the removed nodes. These hierarchical representations of
the shape associated with the global saliency are used to analyze the
shape. Our proposal is different from this framework in terms of the
use of energy minimization. In~\cite{tari.14.jmiv}, the energy
minimization is used to create an image with negative and positive
regions for a given shape. Then one constructs a binary hierarchy of
partitions of the created image via its saddle points, and weighs each
node based on the obtained image values. In our case, the energy
minimization is performed on an input image subordinated to its
hierarchical representation by the tree of shapes. This yields a
quantitative meaningfulness measure {\bluetext
  $\Attribute_{\lambda_s}$} for each node of the tree of shapes.



\section{Conclusion}
\label{sec:conclusion}

In this paper, we have presented an efficient approach of hierarchical
image simplification and segmentation based on minimizing some
multiscale separable energy functional on the tree of shapes, a unique
and equivalent image representation. It relies on the idea of
hierarchy transformation based on extinction values and on a
tree-based shape space to compute a saliency map representing the
final hierarchical image simplification and segmentation.  The salient
structures in images are highlighted in this saliency map. A
simplified image with preservation of salient structures can be
obtained by thresholding the saliency map. Some qualitative
illustrations and quantitative evaluation in context of image
segmentation on a public segmentation dataset demonstrate the
efficiency of the proposed method. A binary executable of the proposed
approach is available on
\url{http://publications.lrde.epita.fr/xu.hierarchymsll}.

In the future, we would like to explore some applications employing a
strongly simplified image as pre-processing step. We believe that this
could be useful for analyzing high-resolution satellite images and
images with texts, where the contours of meaningful objects in images
usually coincide with full level lines. Besides, as advocated in
Table~\ref{tab:singlesegment} for Gpb-owt-ucm, the texture provides
important information for image segmentation. An interesting
perspective is to incorporate texture information in our proposed
framework. Since the tree of shapes is a natural representation of the
input image, a possible way to integrate texture information might
consist in replacing the original image with a new grayscale image
incorporating texture features. Although this is not directly
appicable to our case, probability map incorporating region features,
such as proposed in~\cite{bai2009geodesic}, are worth
exploring. Computing the tree of shapes of such probability maps has
already been proved valuable (see ~\cite{dubrovina.14.icip}). In
another direction, it would be interesting to investigate some other
energy functionals for some specific tasks. Examples are the rate
distortion optimization used in image or video compression coding
system (see~\cite{salembier.00.itip, ballester.07.jmiv}) and the
energy based on spectral unmixing used for hyperspectral image
segmentation in~\cite{veganzones.14.itip}. The energy functionals in
these works are affine separable, which straightforwardly allows to
use them in our proposed framework. Last, but not the least, given
that using a learned gradient improves the results, a major research
avenue is to combine our approach with learning techniques.


\bibliographystyle{abbrv}
\bibliography{xu.2015.prl}

\end{document}